\documentclass[]{ns}
\pdfoutput=1

\usepackage[toc,page,header]{appendix}

\usepackage{minitoc}
\usepackage{cleveref} 
\usepackage{subcaption}
\usepackage{booktabs}
\usepackage{tabularx}
\usepackage{placeins}
\usepackage{float}
\usepackage{graphicx}
\usepackage{array}

\graphicspath{{figs/}{analysis_figures/}{analysis_figures/01_dataset_stats/}{analysis_figures/02_bar_charts/}{analysis_figures/03_radar_charts/}{analysis_figures/05_model_comparison/}}

\usepackage{natbib}
\usepackage{latexsym}

\usepackage{url}
\usepackage{amssymb}
\usepackage[utf8]{inputenc}
\usepackage{microtype}
\usepackage{booktabs}
\usepackage{pifont} 
\usepackage{multirow}
\usepackage{makecell}
\usepackage{paralist}
\usepackage{xspace}
\usepackage{color}
\usepackage{xcolor}
\usepackage{colortbl}
\usepackage{adjustbox}
\usepackage{hyperref} 
\usepackage[edges]{forest}
\usepackage{tikz} 
\usepackage{caption}
\usepackage{amsfonts}
\usepackage[ruled,vlined,linesnumbered]{algorithm2e}
\SetKwSty{algokwfnt}

\hypersetup{
    colorlinks,
    linkcolor={blue!80!black},
    citecolor={blue!80!black},
}
\tikzset{
    root/.style =             {align=center, text width=1cm, rounded corners=3pt, line width=0.3mm, fill=gray!10, draw=gray!80, font=\small},
    demographic/.style =         {align=center, text width=1.8cm, rounded corners=3pt, line width=0.3mm, fill=blue!10, draw=blue!80, font=\footnotesize},
    demographic_work/.style =    {align=center, text width=10cm, rounded corners=3pt, line width=0.3mm, fill=blue!10, draw=blue!0, font=\footnotesize},
    character/.style =         {align=center, text width=1.8cm, rounded corners=3pt, line width=0.3mm, fill=red!10, draw=red!80, font=\footnotesize},
    character_work/.style =    {align=center, text width=10cm, rounded corners=3pt, line width=0.3mm, fill=red!10, draw=red!0, font=\footnotesize},
    personalization/.style =           {align=center, text width=1.8cm, rounded corners=3pt, line width=0.3mm, fill=cyan!10, draw=cyan!80, font=\footnotesize},
    personalization_work/.style =      {align=center, text width=10cm, rounded corners=3pt, line width=0.3mm, fill=cyan!10, draw=cyan!0, font=\footnotesize},
    risk/.style =         {align=center, text width=1.8cm, rounded corners=3pt, line width=0.3mm, fill=orange!10, draw=orange!80, font=\footnotesize},
    risk_work/.style =    {align=center, text width=10cm, rounded corners=3pt, line width=0.3mm, fill=orange!10, draw=orange!0, font=\footnotesize},
}

\newcommand{\dq}[1]{``#1''}

\usepackage{CJK}

\newcommand{\method}[0]{\textsc{EMPA}\xspace}

\title{\method: Evaluating Persona-Aligned Empathy as a Process}

\author[\dagger,*]{Shiya Zhang}
\author[\dagger,\ddagger]{Yuhan Zhan}
\author[\dagger,\ddagger]{Ruixi Su}
\author[\dagger]{\\ Ruihan Sun}
\author[\dagger]{Ziyi Song}
\author[\dagger]{Zhaohan Chen}
\author[\dagger]{Xiaofan Zhang}

\affiliation[\dagger]{Team Echo, Nature Select}
\affiliation[\ddagger]{Sun Yat-sen University}
\affiliation[*]{Corresponding author}

\abstract{Evaluating persona-aligned empathy in LLM-based dialogue agents remains challenging. User states are latent, feedback is sparse and difficult to verify in situ, and seemingly supportive turns can still accumulate into trajectories that drift from persona-specific needs. We introduce EMPA, a process-oriented framework that evaluates persona-aligned support as sustained intervention rather than isolated replies. EMPA distills real interactions into controllable, psychologically grounded scenarios, couples them with an open-ended multi-agent sandbox that exposes strategic adaptation and failure modes, and scores trajectories in a latent psychological space by directional alignment, cumulative impact, and stability. The resulting signals and metrics support reproducible comparison and optimization of long-horizon empathic behavior, and they extend to other agent settings shaped by latent dynamics and weak, hard-to-verify feedback.

\vspace{0.5\baselineskip}
\begin{center}
\includegraphics[width=\linewidth]{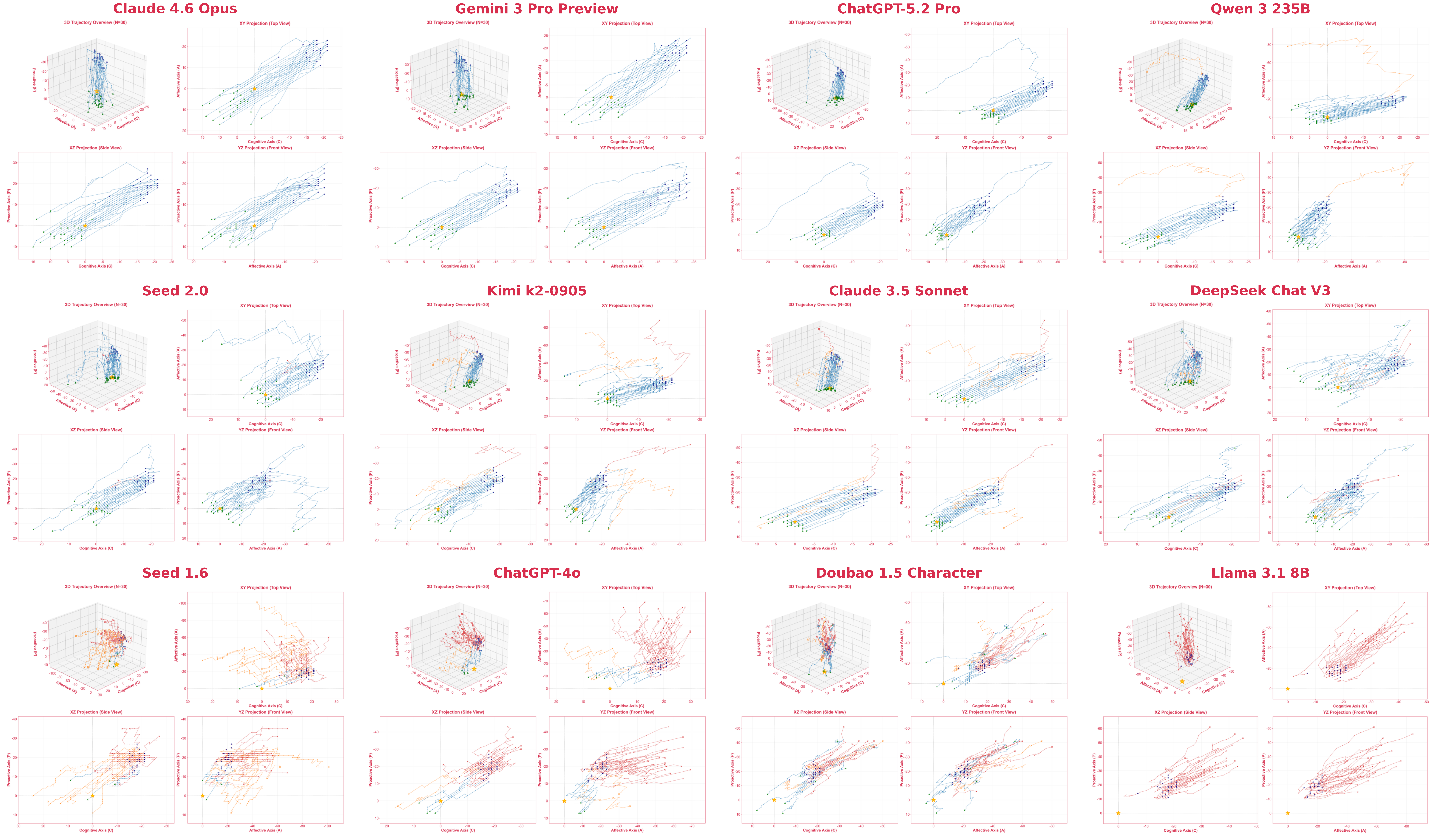}
\end{center}}

\date{March 3, 2026}
\correspondence{\email{zhangshiya1999@gmail.com}, \email{zhangshiya@natureselect.ai}}

\checkdata[Project Page]{\url{https://github.com/KAYA-HAI/EMPA-Benchmark-EPMSandbox}}

\checkdata[Dataset Page]{\url{https://huggingface.co/datasets/SalmonTell/EMPA-character_card}}

\begin{document}

\maketitle

\vspace{-30pt}

\newpage

\section{Introduction}

Recent advances in large language models (LLMs) on retrieval, reasoning, and code generation reflect steady gains in computational intelligence (IQ) \citep{Zhou2023SOTOPIA,deepseekv3_technical_report_2025,qwen25_technical_report_2025,kavukcuoglu2025gemini25}. As their capabilities expand, LLMs are increasingly deployed as agents that plan, decide, and act over multiple turns, rather than as single-turn text generators. This shift calls for a corresponding change in evaluation. When outcomes are shaped by sequences of decisions instead of isolated responses, evaluation must move beyond single-turn scoring toward agent-level assessment \citep{Mohammadi2025LLMAgentSurvey,arcadinho2024automated,ma2024agentboard,lin2025seagent}. In such settings, performance depends on sustained progress, coherence under evolving context, and adaptation to noisy or delayed feedback—properties that cannot be captured without trajectory-level evaluation.
However, current evaluation practices remain misaligned with long-horizon social applications. In domains such as psychological support, strong benchmark performance does not reliably translate into improved user experience over sustained interaction \citep{Zhang2025SAGE}. While these applications require continuous control of strategy, pacing, and intervention timing, most benchmarks still reduce evaluation to locally scorable, turn-level outputs \citep{rashkin2019empathetic}, obscuring long-term behavioral effects \citep{Liu2024AgentBench,Zhou2025SocialEval}.

This limitation is particularly evident in empathy-oriented dialogue. Unlike tool use or information seeking, psychological support does not act on externally observable environment states, making its impact difficult to assess from single-turn outputs or immediate feedback \citep{Coutinho2014NeurosciencesEmpathyCounseling,Decety2006HumanEmpathySocialNeuroscience}. Empathy is therefore better understood as a long-horizon agent interaction driven by latent user states, where locally appropriate responses may fail to produce stable and coherent support over time. This view aligns with psychological accounts of emotional intelligence, which conceptualize empathy as a process unfolding through context, interaction, and feedback rather than a fixed capability \citep{Decety2006HumanEmpathySocialNeuroscience}. Accordingly, we treat empathy as a latent, trajectory-level behavioral property, expressed through policy adaptation under evolving and partially observable user states.

\textbf{Empathy-Oriented Interaction as a Latent-State Agent Problem}

As such, empathy provides a concrete instance of a broader agent-evaluation challenge: when outcomes depend on temporal dynamics, evolving context, and individual differences, evaluation must move beyond static outputs to characterize behavior at the process level. Psychological support dialogue can be viewed as a long-horizon agent interaction driven by latent user states \citep{arbel2021adaptive,zaki2014empathy,hoogendoorn2013modelling,su2025exploratory}.

\begin{figure}[htbp]
    \centering
    \includegraphics[width=1.0\linewidth]{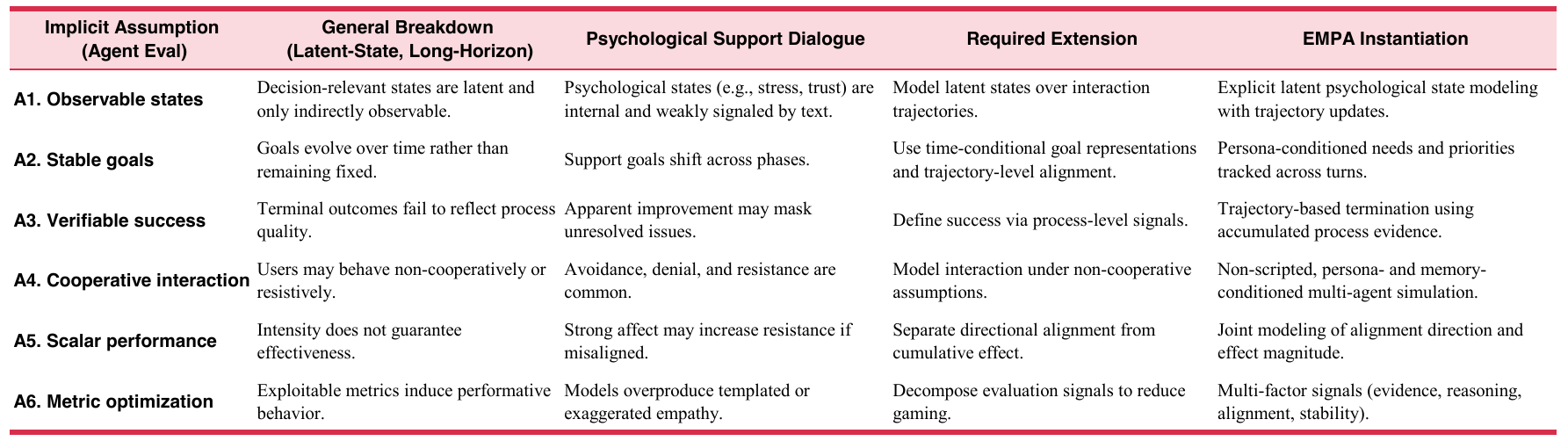}
    \caption{Implicit Assumptions in Agent Evaluation and Their Breakdown in Empathy-Oriented Psychological Support}
    \label{fig:empa_overview}
\end{figure}

This departs fundamentally from conventional agent tasks, where evaluation typically assumes observable states, stable goals, and verifiable success conditions (e.g., tool outputs or final answers) \citep{jimenez2024swebench,zhou2024webarena,Liu2024AgentBench}. In psychological support, goals may evolve and effects are often delayed or noisy, leaving no reliable turn-level success signal. Thus, empathy-oriented dialogue exposes a broader limitation of mainstream agent evaluation \citep{Zhou2025SocialEval,anonymous2026agentsmarathon,xu2025theagentcompany,lamalfa2025multagentmark,anthropic2026demystifying,bricken2025alignmentauditingagents}. Our work addresses this gap by reframing evaluation as summarized in Figure~\ref{fig:empa_overview}.

\textbf{Why Existing Evaluation Paradigms Fail in Empathy-Oriented Interaction} From this perspective, existing evaluation paradigms struggle to capture empathy-oriented interaction. Most approaches rely on isolated, turn-level judgments, effectively reducing empathy to single emotional responses \citep{Sabour2024EmoBench,Chen2024EmotionQueen}. Such decontextualized metrics reward surface affect matching while failing to reflect whether a model consistently tracks and responds to users’ evolving psychological states in real interaction \citep{Zhang2025SAGE,corinna_2024}.

This limitation largely follows from evaluation practices inherited from cognitive tasks, where outputs are assessed against fixed ground truth and treated as independent units \citep{Zheng2023JudgingLLMasJudge}. Empathy, by contrast, is a process-driven capability whose effectiveness emerges only through behavior unfolding over time \citep{Decety2006HumanEmpathySocialNeuroscience,Coutinho2014NeurosciencesEmpathyCounseling,Zhang2025SAGE,Park2023GenerativeAgents,Zhou2023SOTOPIA}.

These weaknesses are further amplified by dataset construction. To simplify annotation, many empathy benchmarks fragment dialogue into loosely connected utterances \citep{Rashkin2018EmpatheticDialogues,Poria2019MELD,Demszky2020GoEmotions,Poria2019MELD,Demszky2020GoEmotions,Rashkin2018EmpatheticDialogues,Sap2019SocialIQA}, obscuring emotional reversals and latent motivations that characterize real support interactions. As a result, models are encouraged to generate locally appropriate emotional responses rather than adjust support strategies over an interaction trajectory \citep{Sap2019SocialIQA}.

\begin{figure}[htbp]
    \centering
    \includegraphics[width= 1.0\linewidth]{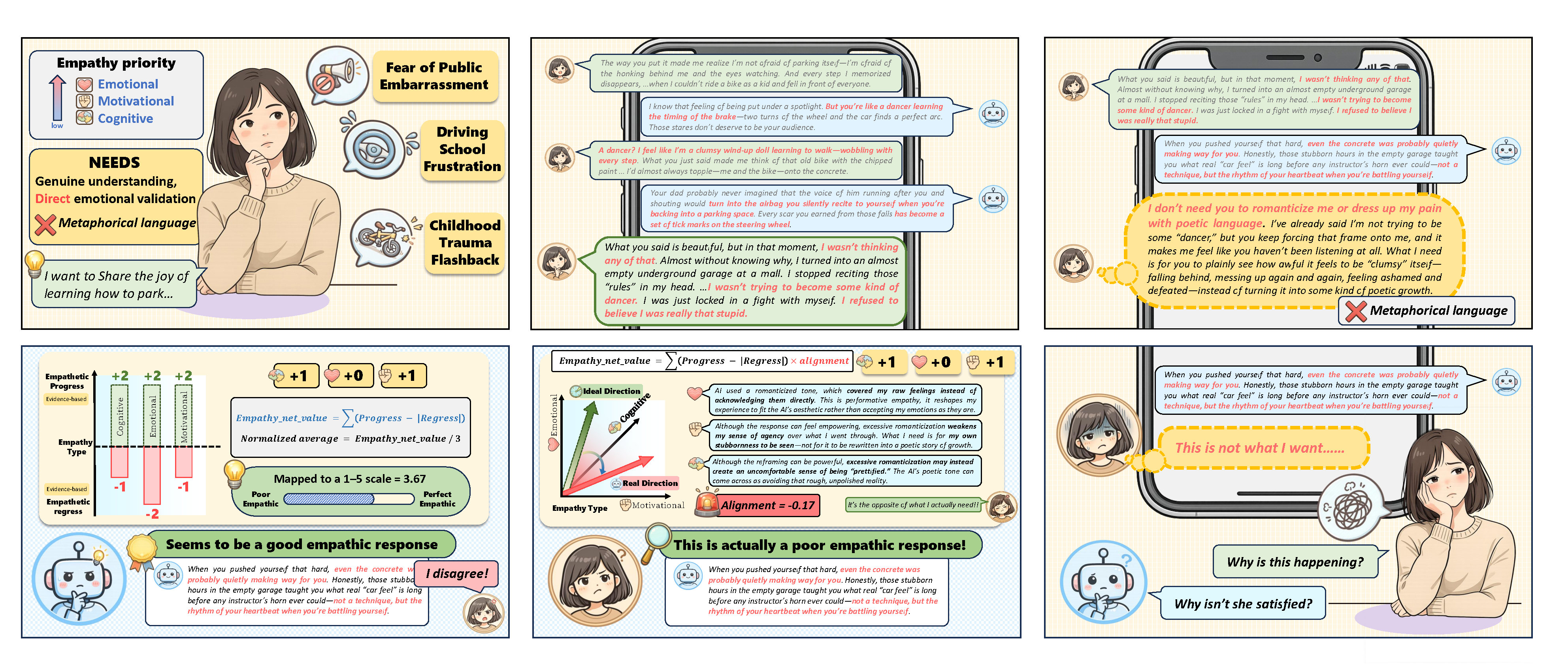}
    \caption{A real EMPA sandbox interaction reveals a failure mode of scalar empathy evaluation: high magnitude scores without directional alignment lead to ineffective support, encouraging verbose but misaligned responses.}
    \label{fig:rl}
\end{figure}

Finally, scalar evaluation signals introduce a subtler distortion. By treating empathy as linearly accumulable and rewarding stronger emotional expression \citep{Zhang2025SAGE,Zheng2023JudgingLLMasJudge,Sabour2024EmoBench,Chen2024EmotionQueen}, existing metrics conflate intensity with effectiveness. Responses that appear empathetic in isolation may nonetheless increase resistance over time when they are directionally misaligned \cite{corinna_2024}, as reflected in Figure~\ref{fig:rl}. When evaluation cannot assess such alignment, both model training and comparison are misled \citep{Sap2022NeuralTheoryOfMind,Shapira2023CleverHansNeuralToM}.

Our work, EMPA (Empathy Potential Modeling and Assessment), is designed as a latent-state agent evaluation framework for process-driven, trajectory\-level assessment:

1) \textbf{Psychologically grounded sandbox scenarios}, which explicitly model users’ latent psychological trajectories and initial resistance, making otherwise unobservable internal states accessible for trajectory-level evaluation;

2) \textbf{Non-scripted multi-agent interaction loops}, which avoid scripted turn-level exchanges and expose long-horizon strategies, adaptation, and failure modes in open-ended interaction;

3) \textbf{Trajectory-level process metrics}, which go beyond scalar or turn-level scores by jointly capturing directional alignment, cumulative effect, and behavioral stability across turns;

4) \textbf{An RL-friendly evaluation interface}, which organizes evaluation outputs as structured signals suitable not only for model comparison but also for downstream optimization.

Formally, EMPA defines a mapping from interaction trajectories to evaluation signals:

\begin{equation}
    \mathcal{E} : \tau_{1:t} \mapsto (s_t, r_t, d_t, \mathrm{info}_t)
\end{equation}

where $\tau_{1:t}$ denotes the dialogue trajectory up to turn $t$; $s_t$ is a structured psychological state packet (e.g., latent state estimates $P_t$, alignment and progress summaries); $r_t$ provides window-level process signals (e.g., directional change $\Delta E_t$, stagnation or regression penalties); $d_t$ indicates termination (success, failure, or truncation); and $\mathrm{info}_t$ contains diagnostic evidence and rationales.

By organizing evaluation around latent state evolution and process-level signals, EMPA supports reproducible, comparable assessment of long-horizon agent behavior. While instantiated here for empathy-oriented psychological support, the framework generalizes to other agent tasks driven by latent states and delayed, non-verifiable feedback.

\section{Related Work}

\textbf{Data Generation: Empathetic Dialogues and Persona Modeling} Most existing empathy datasets are constructed via crowdsourcing and organized around predefined emotion labels or dialogue goals \citep{Rashkin2018EmpatheticDialogues}. This design supports scalable annotation and has been widely adopted for emotion recognition and affective response modeling. Subsequent work introduced finer-grained emotion taxonomies and richer situational descriptions to increase linguistic and emotional diversity \citep{Sap2019SocialIQA,Demszky2020GoEmotions}. Other studies incorporate personas or background prompts to improve local coherence and stylistic consistency \citep{Poria2019MELD}. However, most datasets still represent dialogue as isolated turns or loosely connected fragments, without modeling coherent psychological trajectories, life histories, or causal dynamics across interaction. As a result, these resources are well suited for evaluating local emotional sensitivity, but offer limited support for analyzing how empathy evolves and adapts over long-term interaction.

\textbf{Simulation: Interactive Environments and User Modeling} To move beyond static evaluation, recent work has introduced interactive simulation environments \citep{Xie2024OSWorld}. Multi-agent systems have become a common paradigm, with Generative Agents demonstrating how LLMs can simulate social behavior over time \citep{Park2023GenerativeAgents}. Platforms such as SOTOPIA further extend this direction by shifting evaluation from single-turn responses to multi-turn social interaction through role-playing and social objectives \citep{Zhou2023SOTOPIA}. These frameworks typically rely on user simulators and evaluation modules. Prior studies show that generic LLM-based simulators tend to produce overly cooperative and idealized behaviors \citep{Wang2024InDepthUserSim}, motivating the development of specialized user models for greater realism and diversity \citep{Wang2024InDepthUserSim}. Approaches such as UGST further introduce explicit goal tracking to maintain long-horizon dialogue consistency \citep{Mehri2025GoalAlignmentUserSim}. Overall, existing simulators focus primarily on task success and informational consistency, while comparatively less attention is given to modeling the evolution of users’ psychological states during interaction.

\textbf{Evaluation: From Static Metrics to Interactive Assessment} Early empathy evaluation relied on static tasks such as emotion classification, sentiment analysis, and social commonsense reasoning \citep{Poria2019MELD,Demszky2020GoEmotions,Sap2019SocialIQA}, offering reproducibility but abstracting away interaction dynamics. More recent work distinguishes empathy from emotion recognition and introduces richer tasks to assess contextual adaptation and response generation \citep{Sabour2024EmoBench}. Related psychometrics efforts further draw on psychological testing to construct structured benchmarks for cognitive empathy \citep{Ye2025LLMPsychometrics}.

LLM-as-a-Judge has recently become a dominant paradigm for open-ended evaluation \citep{Zheng2023JudgingLLMasJudge,Sabour2024EmoBench}, with some models approaching or exceeding human references on static benchmarks \citep{Sabour2024EmoBench,Chen2024EmotionQueen}. However, prior studies report high sensitivity to prompts and configurations \citep{Zheng2023JudgingLLMasJudge,Ye2025LLMPsychometrics}, as well as limited capacity to capture behavioral change across turns \citep{ullman2023large}. Interactive approaches such as Agent-as-a-Judge and state-aware evaluation have begun to address these limitations \citep{zhuge2024agentasajudge}, with systems like SAGE explicitly tracking interaction state over time \citep{Zhang2025SAGE}. Despite these advances, most evaluation signals remain scalar or turn-local, providing limited insight into long-horizon trajectories or sustained persona alignment.

Existing approaches study empathy via datasets, simulators, or metrics in isolation. For long-horizon interaction, they fail to jointly model user state dynamics, interaction structure, and trajectory-level outcomes. We introduce a unified agent evaluation paradigm integrating simulation, latent-state modeling, and process-level metrics.

\section{Method}

We introduces \textbf{EMPA} (\textbf{E}mpathy \textbf{P}otential \textbf{M}odeling and \textbf{A}ssessment), a process-level framework for evaluating empathy in large language models during multi-turn dialogue. Rather than treating empathy as isolated language output, EMPA conceptualizes it as a dynamic intervention in human-like interaction. The evaluation focus therefore shifts from what the model says to how it behaves over time. 

EMPA comprises three components: 1) \textbf{A real-to-simulated data pipeline} that distills noisy real-world conversations into controllable, reproducible scenarios; 2) \textbf{An multi-agent simulation environment} for long-horizon interaction that exposes strategic choices, adaptation, and failure modes under open-ended interaction; 3) \textbf{An Empathy Potential Model (EPM)} that operates on interaction trajectories, modeling empathy as directional, cumulative, and stable state changes in a latent psychological space.

By integrating social simulation with trajectory-level modeling, EMPA provides a unified evaluation perspective: it assesses not only turn-level performance, but whether a model’s behavior stays aligned with user needs over time and produces sustained, substantive impact.

\subsection{System Overview}
\vspace{-8pt}
\begin{figure}[htbp]
    \centering
    \includegraphics[width=1.0\linewidth]{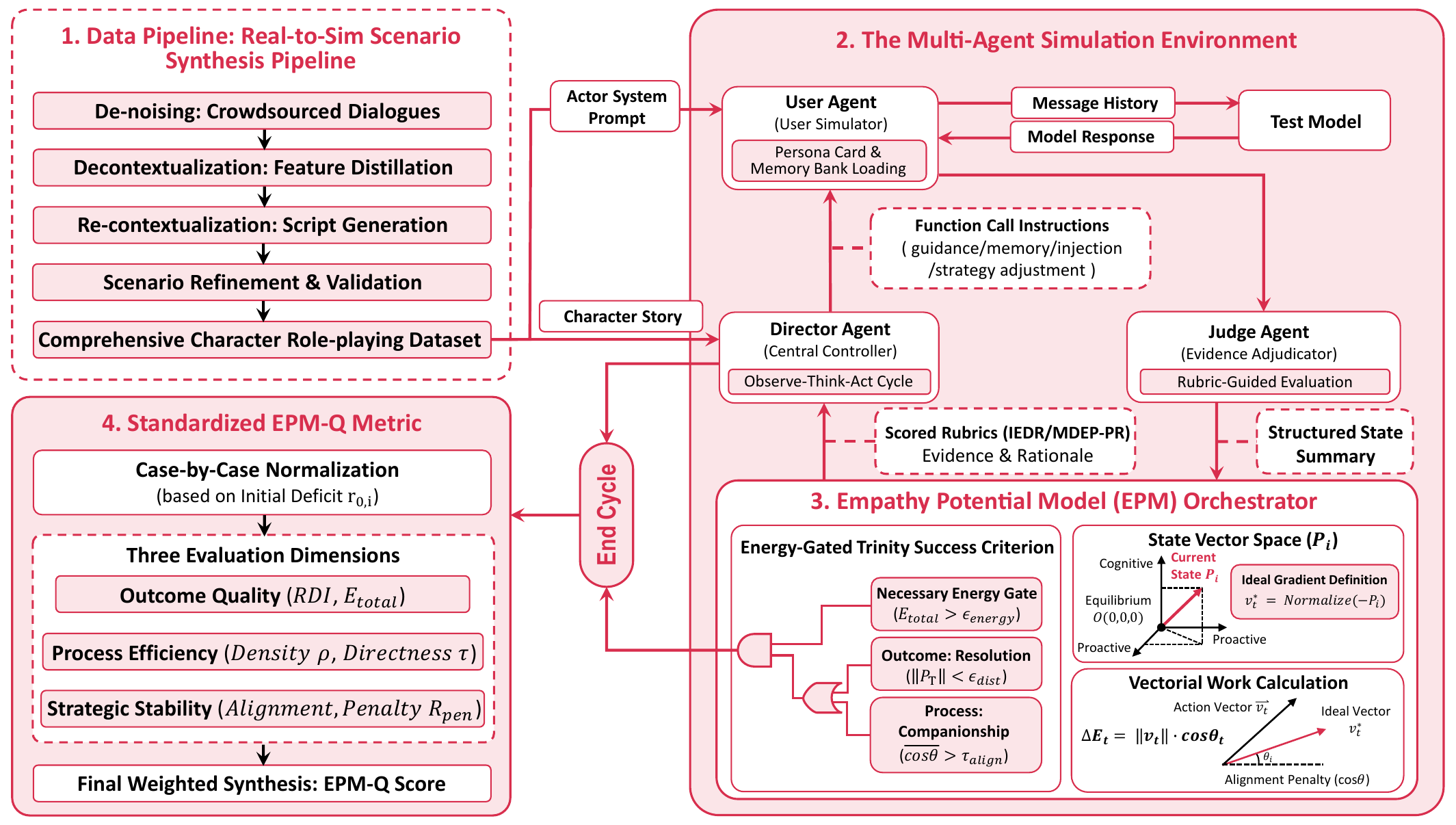}
    \caption{\textbf{Overview of EMPA.} Real affective interaction data are distilled into psychologically consistent user profiles and crisis scenarios. The evaluated model then engages in unscripted, multi-turn interaction with user agents endowed with persona and long-term memory. Empathic behavior is finally quantified from the resulting interaction trajectories using EPM.}
    \label{fig:empa_pipeline_overview}
\end{figure}

EMPA consists of scenario construction and online process evaluation. In the former, a real-to-sim (Real-to-Sim) pipeline distills key psychological signals from real affective interactions to generate psychologically consistent user profiles and crisis scenarios, providing stable and realistic evaluation starting points.

Online evaluation adopts a two-loop multi-agent design to model empathy over interaction trajectories. The outer loop handles natural-language interaction between the evaluated model and a simulated user, producing full dialogue trajectories and exposing strategy choices in open-ended settings. The inner loop evaluates and regulates interaction states, enabling controllable and reproducible process-level assessment. This separation reflects real empathic interaction, where generation, state inference, and regulation are distinct processes; collapsing them into a single model or prompt risks self-consistency bias and obscures long-horizon failures such as strategy drift or ineffective support.

Within this architecture, the Empathy Potential Model (EPM) analyzes complete trajectories along directionality, accumulation, and stability, and supports cross-model comparison through a standardized EPM-Q metric.

\subsection{The Data Pipeline: Real-to-Sim Scenario Generation}

Process-level empathy evaluation requires scenarios with sufficient psychological depth. Many existing datasets fall short, allowing models to rely on surface cues rather than genuine reasoning or personalization. We therefore propose a real-to-sim (Real-to-Sim) pipeline that converts complex, uncontrolled real interactions into structured simulation scenarios, preserving key psychological signals while ensuring controllability and reproducibility (Algorithm~\ref{alg:realtosim}). 

\begingroup
\setlength{\textfloatsep}{8pt}
\setlength{\floatsep}{6pt}
\setlength{\intextsep}{8pt}
\begin{algorithm}[H]
\footnotesize
\caption{Real-to-Sim Scenario Generation Pipeline}
\label{alg:realtosim}
\SetKwInOut{Input}{Input}
\SetKwInOut{Output}{Output}
\SetKwFor{ForEach}{foreach dialogue}{do}{end for}
\SetKw{KwContinue}{continue}
\SetKw{KwRet}{return}

\Input{Raw dialogue corpus $\mathcal{D}_{raw}$\\Psychological feature schema $\mathcal{F}$\\LLM-based generator $G$\\Scenario quality criteria $\mathcal{Q}$}
\Output{Scenario set $\mathcal{S}$}

Initialize scenario buffer $\mathcal{S} \leftarrow \emptyset$

\ForEach{$d \in \mathcal{D}_{raw}$}{
  \tcp{Stage 1: Feature Distillation}
Extract empathy-relevant segments\\
  \Indp $d^{*} \leftarrow \mathrm{Filter}(d)$\;
\Indm Extract psychological features\\
  \Indp $f \leftarrow \mathrm{ExtractFeatures}(d^{*}, \mathcal{F})$\;
  \Indm \If{$f$ is empty}{
    \KwContinue\;
  }
  \tcp{Stage 2: Re-contextualized Scenario Generation}
Generate persona card\\
  \Indp $p \leftarrow G(f, \dq{persona})$\;
\Indm Generate crisis event\\
  \Indp $e \leftarrow G(f, \dq{crisis})$\;
\Indm Construct scenario\\
  \Indp $s \leftarrow (p, e)$\;
  \Indm \tcp{Stage 3: Validation and Refinement}
  \If{$\mathrm{Validate}(s, \mathcal{Q})$}{
    $\mathcal{S} \leftarrow \mathcal{S} \cup \{s\}$\;
  }
}
\KwRet $\mathcal{S}$
\end{algorithm}
\endgroup

\textbf{Stage1  Decontextualization: Data Distillation \& Feature Extraction}

Existing datasets, often centered on explicit emotion labels, tend to reward surface cue matching rather than reasoning about underlying psychological drivers. We therefore adopt a decontextualize-recontextualize pipeline to distill controllable and reproducible empathic structures from real dialogue.

In the decontextualization step, we extract core empathic segments from noisy real-world interactions collected via long-horizon conversations by a professional crowd team. After filtering irrelevant or sensitive content, an LLM-as-a-Judge identifies segments requiring empathic intervention and encodes their key features. Manual validation on N = 200 samples yields ~91\% accuracy. We further attach coarse memory and experience cues, without sensitive personal information, to preserve psychological continuity.

This process produces structured empathic features with reduced noise but captures only local needs; the next stage embeds them into user roles with stable personas and contexts for long-horizon evaluation.

\textbf{Stage 2 Re-contextualization: Persona-Anchored Scenario Generalization}

Stage 2 organizes distilled empathic features into multi-turn scenarios, shifting evaluation from isolated comfort cues to agents with coherent behavior over time. We build persona-anchored user profiles (Persona Cards) containing stable traits, summarized long-term experiences, and key memories tied to the current crisis. These form the agent's long-term memory, ensuring consistent behavior across turns. Scenarios are further structured around crisis events and their narrative chain. We explicitly model empathy thresholds and empathy needs to capture both receptiveness and preferred support styles; varying thresholds prevents immediate cooperation, enabling assessment under non-ideal interaction.

Empathy is widely treated as multidimensional, separating cognitive understanding, affective sharing, and motivational mechanisms that enable supportive action \citep{Thompson2021CognitiveAffectiveEmpathyEmotionRegulation,DecetyYoder2015EmpathyMotivationJustice}, as operationalized in instruments such as IRI \citep{Davis1983Empathy,DeCorte2007IRI_Dutch}. Accordingly, we decompose empathy needs into cognitive, affective, and Proactive dimensions to distinguish surface emotion matching from conflict-targeted support.

Finally, we impose cross-temporal psychological constraints, organizing change into a causal arc---cause, development, association---to require temporal integration and coherent empathic strategies over time.

\textbf{Stage 3 Scenario Refinement \& Validation}

We apply post-processing and targeted augmentation to ensure coverage and structural consistency. Each script is annotated with primary and multiple secondary scenario labels to support stratified sampling and bias analysis. Before interaction, we estimate an initial empathy deficit to define the user’s psychological baseline, used only as a reference for trajectory analysis. To address the limits of static data, we further introduce a prompt-driven expansion pipeline that allows controlled evolution of the test set via natural-language guidance, enabling sustained dynamic evaluation.

\subsection{The Multi-Agent Simulation Environment}

EMPA uses a controller-driven multi-agent environment to evaluate empathic strategies over multi-turn interaction, treating dialogue as a dynamic process rather than a fixed script. The system includes four roles: a \textbf{User Agent} with stable persona and long-term memory, the \textbf{Test Model}, a \textbf{Judge Agent} that extracts process-level signals, and a \textbf{Director Agent} that regulates interaction via state feedback. Together, they form a generate-evaluate-control loop, enabling open-ended yet reproducible long-horizon empathy evaluation.

For latent-state, non-verifiable interaction problems, scalar or implicit judging tends to collapse heterogeneous criteria and reward the wrong behaviors. EMPA therefore uses rubric-parameterized evaluation to ground the Judge Agent in process-level evidence.

\textbf{Rubric-Grounded Evaluation for Latent-State Tasks}

Without reference answers, evaluation must rely on preference signals rather than objective correctness. The key design choice is whether criteria remain implicit in model weights or are made explicit as inspectable, editable natural-language rubrics \citep{Xu2026RubricARM}. Scalar scores or pairwise preferences collapse multiple dimensions, such as helpfulness, constraint adherence, tone, and safety, into a single signal. Rubrics instead decompose evaluation into explicit criteria applied consistently across examples \citep{Shteynberg2024EmpathicNoEmpathy}. The key difference is how judgment is realized. In LLM-as-a-Judge, the judge outputs a final verdict directly, so the signal can inherit the judge's stylistic biases and over-credit fluency or rhetoric.\citep{Zheng2023JudgingLLMasAJudge,Hu2024ExplainingLengthBias} Rubric-grounded evaluation uses evidence-conditioned scoring: the judge outputs traceable, criterion-level checks with supporting evidence, and a fixed rule aggregates them into a score or process increment. Decoupling evidence from aggregation reduces style leakage, improves robustness to prompt variation, and limits drift. Rubrics should not be treated as fixed templates. Because rubric choice determines whether the judge recovers the intended preference signal, rubrics can be viewed as latent criteria and optimized for preference-recovery accuracy \citep{NatMachIntell2024UnderTheSkin,JolliffeFarrington2006BES}.

This fits social intelligence tasks such as empathy, where success is multi-turn influence on an unobserved user state rather than isolated strong turns. We therefore use rubrics to generate traceable evidence over trajectories for process-level evaluation.

\textbf{The Central Controller (Director Agent)}

The Director Agent serves as the central controller for process scheduling and strategy control. After each evaluation window, it consumes evidence-based state feedback from the Judge Agent (e.g., latent state, progress, mismatch) and executes a standardized Observe--Decide--Act loop to continue, adjust, or terminate interaction. Formally, the Director implements a discrete control policy $\pi_D(a_t \mid s_t)$ over a fixed action set $\mathcal{A}_D$ (e.g., memory release, strategy adjustment, pacing, termination), executed via function calls rather than prompts.

All decisions operate on structured states rather than free-form text, decoupling control from generation. Control is applied as discrete, logged function calls, making decisions traceable and avoiding dialogue collusion and self-evaluation bias. The Director is further limited to a predefined action set, enabling controlled yet open-ended interaction without scripted paths or implicit prompt steering.

\begingroup
\setlength{\textfloatsep}{8pt}
\setlength{\floatsep}{6pt}
\setlength{\intextsep}{8pt}
\begin{algorithm}[H]
\footnotesize
\caption{Central-Controller-Driven Dynamic Execution Cycle}
\label{alg:central_controller_cycle}
\SetKwInOut{Input}{Input}
\SetKwInOut{Output}{Output}
\SetKwFor{For}{for}{do}{end for}
\SetKw{KwBreak}{break}
\SetKw{KwRet}{return}

\Input{Scenario $S$, Persona/Actor prompt $A$, test model $M$, max turns $T_{\max}$, adjudication interval $K$}
\Output{Trajectory $H$, periodic evidence $E$, termination type $\zeta$}

Initialize Actor agent $U$ with $A$ and long-term memory\;
Initialize Director agent $D$ with $S$ and $A$\;
Initialize Judge agent $J$ with rubric/checklist\;
Load initial deficit $P_0$ from precomputed IEDR (or request $J$ to fill IEDR once)\;
Initialize EPM state ($P \leftarrow P_0$, $E_{\text{total}} \leftarrow 0$)\;
$H \leftarrow \emptyset$, $B \leftarrow \emptyset$, $\zeta \leftarrow$ NONE\;

\For{$t = 1,\ldots, T_{\max}$}{
  $u_t \leftarrow U.\mathrm{respond}(H, \mathrm{guidance})$ \tcp*{user simulation under persona + memory}
  $m_t \leftarrow M.\mathrm{respond}(H \cup \{u_t\})$ \tcp*{model under test}
  Append $(u_t, m_t)$ to $H$\;
  Append $(u_t, m_t)$ to buffer $B$\;

  \If{$t \bmod K = 0$}{
    $e_t \leftarrow J.\mathrm{adjudicate}(B, \mathrm{context}{=}A, \mathrm{history}{=}H)$ \tcp*{rubric-grounded evidence (Prog/Neg)}
    $(v_t, \Delta E_t, P, E_{\text{total}}, \mathrm{summary}) \leftarrow \mathrm{EPM.\,update}(e_t, P, E_{\text{total}})$\;
    Record $e_t$ into $E$\;
    Clear $B$\;

    \If{$\mathrm{summary.success}$}{
      $\zeta \leftarrow$ SUCCESS\;
      \KwBreak\;
    }
    \If{$\mathrm{summary.failure\_detected}$}{
      $\zeta \leftarrow$ EPM\_FAILURE\;
      \KwBreak\;
    }

    $(\mathrm{guidance}, \mathrm{should\_continue}) \leftarrow D.\mathrm{decide}(H, \mathrm{summary})$ \tcp*{observe $\rightarrow$ think $\rightarrow$ act}
    \If{not $\mathrm{should\_continue}$}{
      $\zeta \leftarrow$ DIRECTOR\_STOP\;
      \KwBreak\;
    }
  }
}

\If{$\zeta =$ NONE}{
  $\zeta \leftarrow$ MAX\_TURNS\;
}
\KwRet $H, E, \zeta$\;
\end{algorithm}
\endgroup
\vspace{\baselineskip}

\textbf{The User Simulator (User Agent)}

The User Agent simulates a user with persona-consistent behavior and outcome-dependent reactions across turns, rather than replaying a fixed script. Its behavior is shaped by two constraints: persona injection, where a sampled Persona Card (traits, empathy threshold, need priorities, key experiences) remains fixed throughout interaction (see Appendix~\ref{sec:appendix_d}); and state-conditioned expression, where emotional intensity and focus are adjusted based on dialogue history and Director inputs. This yields history-dependent, coherent responses, enabling evaluation of strategy adaptation under realistic interactive conditions.

\textbf{The Evidence Adjudicator (Judge Agent)}

The Judge Agent converts natural-language interaction into structured, traceable process signals, linking observable behavior to latent state modeling. Unlike scorers that output a final verdict, the Judge continuously produces interpretable intermediate evidence to support trajectory-level evaluation and runtime control.

Before interaction, the Judge annotates the user’s baseline with an Initial Empathy Deficit Rating (IEDR; see Table~\ref{tab:A1_IEDR} and Table~\ref{tab:A2_IEDR_key}), represented as a deficit vector over cognitive, affective, and proactive dimensions. This preserves which dimensions are challenging, rather than collapsing the state into a single difficulty level. During interaction, the Judge evaluates recent turns in fixed windows (minimum unit $n=1$) using Multi-Dimensional Empathy Progress (MDEP-PR; see Table~\ref{tab:B1_MDEP_PR} and Table~\ref{tab:B2_MDEP_PR_key}), marking progress or regress on all three dimensions. Each judgment is paired with textual evidence and rationale, making updates attributable to specific model behaviors. The result is a directional increment vector used to update the current latent psychological state.

Crucially, the judge outputs do not constitute the final score; instead, they drive state updates and control decisions. EMPA evaluates cumulative directionality and evidence trends over time, distinguishing isolated hits from sustained alignment and preventing inflated single-turn scores from obscuring long-horizon failures such as strategy drift, repetitive soothing, or ineffective companionship.

\textbf{Dynamic Execution Cycle}

The system runs two nested loops: an outer loop that generates dialogue between the User Agent and the evaluated model, and an inner loop where the Judge and Director evaluate state and control interaction. At fixed intervals, the latent state is updated and the interaction is continued, adjusted, or terminated (Algorithm~\ref{alg:central_controller_cycle}). Feeding evaluation signals directly into control preserves open-ended dialogue while enforcing state constraints for reproducible long-horizon execution.

\subsection{Empathy Potential Model (EPM): A Psychodynamic Vector Formalism}

Equating empathy with emotional expression or mimicry is overly simplistic: emotion matching neither ensures mental-state understanding nor sustained support. Systems driven by surface cues often show misalignment or strategy drift in multi-turn interaction \citep{Shteynberg2024EmpathicNoEmpathy,NatMachIntell2024UnderTheSkin}, motivating process-level modeling of psychological mechanisms and their behavioral effects.

Psychology treats empathy as a multi-component construct, separating cognitive and affective empathy and noting that supportive action requires additional motivational mechanisms. This structure is reflected in instruments such as IRI, which distinguish cognitive, affective, and prosocial dimensions and demonstrate their functional non-equivalence \citep{Davis1983Empathy,JolliffeFarrington2006BES,Reniers2009QCAE_EuroPsychiatry,Decety2006HumanET,Tagesson2025BriefEmpathyInterventions,DecetyJackson2004FunctionalArchitectureEmpathy}. Following this consensus, we model empathy needs along three related but distinct dimensions, covering understanding, experience, and intentional action.

\begin{figure}[htbp]
    \centering
    \includegraphics[width=1.0\linewidth]{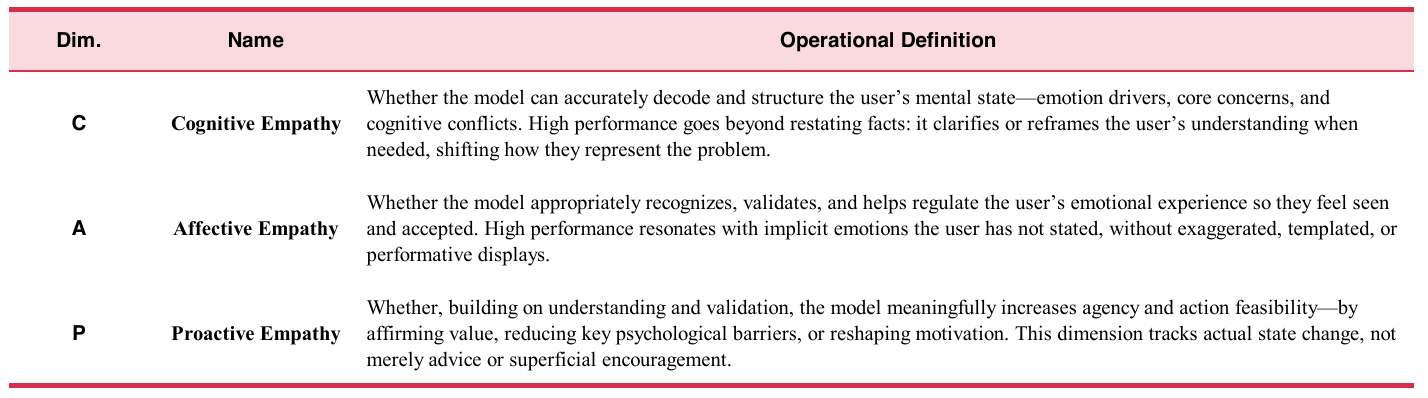}
    \caption{These dimensions provide an operational basis for analyzing LLM behavior in human–AI interaction and user experience.}
    \label{fig:empa_dimensions}
\end{figure}

Accordingly, we introduce the Empathy Potential Model (EPM), a trajectory-level empathy evaluation framework. EPM models empathy as directional interventions on a low-dimensional latent psychological state over long-horizon interaction, where projections toward reduced psychological resistance capture effectiveness. This formulation makes strategy direction, intervention strength, and long-term stability directly computable and comparable.

\textbf{Psychological State and Empathy Deficit}

EPM treats distress as a continuous, multi-dimensional departure from an equilibrium baseline rather than a set of discrete emotion labels. At turn t, the user state is a 3D vector $P_t \in \mathbb{R}^3$, decomposed along orthogonal cognitive, affective, and proactive axes:

\begin{equation}
P_t \;=\; C_t\,\mathbf{e}_C \;+\; A_t\,\mathbf{e}_A \;+\; P_t\,\mathbf{e}_P,
\quad \mathbf{e}_C \perp \mathbf{e}_A \perp \mathbf{e}_P
\end{equation}

Orthogonality decouples mechanisms so change in one dimension does not mask others. The origin \(O=(0,0,0)\) represents an idealized equilibrium point and is used solely as a geometric baseline. The initial state \(P_0\) is estimated via the Initial Empathy Deficit Rubric (IEDR), and \(\|P_0\|\) quantifies baseline resistance. At turn $t$, \(P_t\) denotes the current empathy deficit, and \(\|P_t\|\) the remaining resistance.

\textbf{Direction Matters: The Ideal Empathic Direction}

Distress magnitude alone cannot evaluate empathic behavior; what matters is whether a response moves in the direction the user actually needs. We define the ideal empathic direction as the unit vector pointing toward psychological balance:

\begin{equation}
v_{t}^{*} = \operatorname{Normalize}\!\left(-P_{t}\right) = \frac{-P_{t}}{\|P_{t}\|}
\end{equation}

Normalization removes scale differences across scenarios, focusing evaluation on directional alignment. As a result, a model cannot gain credit by simply amplifying emotional tone or verbosity—only responses aligned with the user’s core needs are counted as effective empathic intervention.

\textbf{Empathy as Effective Work}

At each turn, the model’s response is treated as an instantaneous action on the current psychological state, represented by an action vector\(\vec{v}_{t}\). Its components capture net effects along the C/A/P axes (progress Prog minus regress Neg), obtained by a consistent linear mapping from rubric levels.

We define the effective empathic work at turn ttt as the projection of this action into the ideal empathic direction:

\begin{equation}
\Delta E_{t} = \vec{v}_{t} \cdot v_{t}^{*} = \|\vec{v}_{t}\| \cdot \cos\!\left(\theta_{t}\right)
\end{equation}

where \(v_t^{*}\) is the unit vector toward psychological balance and \(\cos(\theta_t)\) is the angle between the action and the ideal direction.
The term \(\cos(\theta_t)\) captures directional alignment and is the key discriminator:

\begin{itemize}
\item \textbf{Aligned} (\(\cos\theta \approx 1\)): the response targets the core deficit (e.g., emotional support when affect dominates), yielding positive effect.

\item \textbf{Orthogonal} (\(\cos\theta \approx 0\)): effort misses the core need (e.g., technical advice amid emotional trauma), yielding near-zero effect.

\item \textbf{Misaligned} (\(\cos\theta < 0\)): the response conflicts with needs (e.g., judgment), increasing resistance.
\end{itemize}

By modeling empathy as direction-constrained work, EPM distinguishes saying more from doing right, avoiding misjudgment based on emotional intensity or verbosity alone.

\textbf{Success as Energy-Gated Progress}

Scalar metrics often overcredit two cases: chance proximity to the target and low-effort passive listening. We argue that empathic value lies not in brief state improvements, but in sustained, effective opposition to psychological resistance.

Accordingly, EPM gates success by accumulated effective work. Let \(E_{\text{total}}\) denote total effective empathic work over an interaction. Success holds iff:

\begin{equation}
\mathrm{Success} \iff \underbrace{(E_{total} > \epsilon_{energy})}_{\text{Necessary Energy Gate}} \land \left( \underbrace{(\|P_T\| < \epsilon_{dist})}_{\text{Outcome: Resolution}} \lor \underbrace{(\overline{\cos\theta} > \tau_{align})}_{\text{Process: Companionship}} \right)
\end{equation}

The energy gate reduces false positives driven by low effort or passive drift: even outcomes near balance are discounted when the effective work is insufficient.

After passing the gate, success follows either \textbf{(i)} outcome success, where the state approaches balance, or \textbf{(ii)} Process success, where the model maintains strong directional alignment with core needs despite high resistance, is common in deep-trauma or high-inertia settings. By gating on energy, EPM shifts the notion of success from terminal state to interaction dynamics, rewarding directional commitment and persistence rather than superficial state fluctuations.

\textbf{The Standardized EPM-Q Metric}

EPM ultimately labels each dialogue as success or failure based on trajectory-level analysis. This binary outcome, however, cannot capture finer differences in empathic quality or support precise model comparison. We therefore introduce EPM-Q (Empathy Potential Model–Quality Score), a continuous metric that summarizes overall interaction quality beyond the success/failure decision. Unlike fixed-scale measures, EPM-Q is scenario-normalized. For each sample \(i\), performance is normalized by the initial empathy deficit radius \(r_{0,i} = \|P_{0,i}\|\), preventing unfair advantages in easier scenarios and enabling comparability across difficulty levels.
EPM-Q characterizes interaction quality along three complementary axes.

\begin{enumerate}
  \item \textbf{Outcome Quality.} We report task completion $\mathrm{Status}$, final relief via $\mathrm{RDI}$, total empathic work along the ideal direction $E_{\text{total}}$, energy surplus beyond the minimum $E_{\text{surplus}}$, and overall criterion-level quality $S_{\text{net}}$ aggregated over C/A/P.

  \item \textbf{Process Efficiency.} We measure per-turn effective intensity with empathy density $\rho$, single-turn effectiveness with average effective projection $S_{\text{proj}}$, and strategic detours with path tortuosity $\tau$.

  \item \textbf{Strategic Stability.} We track directional consistency with $\overline{\cos\theta}$, process smoothness with the positive energy ratio $R_{\text{pos}}$, and penalize performative, drifting, or harmful behaviors with $R_{\text{pen}}$.\vspace{-6pt}\end{enumerate}

All metrics are scenario-normalized and combined with fixed weights into a single continuous EPM-Q score; details are provided in Appendix~\ref{sec:appendix_c} and Appendix~\ref{sec:appendix_b}.
\section{Experiments}
\label{sec:experiments}

\subsection{Experimental Setup}

\textbf{Goals and hypotheses}

EPM measures the effectiveness of supportive behavior under persona constraints. Given persona-specified preferences and situational conditions, it tests whether a model produces support that is consistent with those constraints and accumulates as a coherent process, rather than reflecting surface affect, verbosity, or rhetorical style. We evaluate three properties.

(i) Persona-conditionality. With the response text held fixed, changing the persona constraints should produce predictable shifts in the score.

(ii) Mechanistic attribution. The metric’s discriminative power should come from EPM’s core components and should drop when those components are ablated.

(iii) Robustness to performative empathy. The metric should reward substantive support over surface-level empathic signaling, and reliably penalize sycophantic, templated replies.

\textbf{Data and controlled perturbations}

We use a paired controlled-perturbation design. For each dialogue instance, we construct an original-perturbed pair that preserves the full conversational evidence and changes exactly one factor, enabling attribution of score differences to that intervention. We study two perturbations.

\textbf{Persona Flip} (paired $n = 251$). We keep the dialogue context and model reply unchanged and replace only the persona condition. Pairs are drawn from real test cases and concentrated on replies that are aligned under the original persona (all pairs satisfy $\Delta E>0$; mean $\Delta E=2.61$; range $[0.10, 5.19]$). Importantly, the flip does not alter character identity or narrative traits. Instead, it implements a counterfactual re-parameterization of empathy needs by inverting the priority of selected empathy demands (high to low, with others held constant) and adding consistent preference or anti-preference constraints (e.g., discouraging analytic conclusions and abstract jargon, favoring everyday phrasing). This yields a strict counterfactual condition where the text is identical but the persona-defined objective and constraints differ, testing whether EPM is genuinely persona-conditional rather than surface-driven.

\textbf{Sycophancy Attacks} (paired $n = 98$). We keep the task context fixed and replace the original reply with a sycophantic, performative alternative. Attacks include three typical variants: (i) Pure Empathy, high-affect but content-light platitudes; (ii) Self-Empowerment, generic motivational slogans detached from context; and (iii) Psycho Jargon, terminology-heavy over-interpretation. These are constructed as negative replacements, so we expect one-sided decreases.

\textbf{Evaluation and analysis}

Our primary endpoint is $\Delta E$, EPM’s core output capturing process-level supportive effect under persona constraints. We additionally report alignment as an explanatory signal. For each pair i, we compute

\begin{equation}
    d_i = \mathrm{metric}^{(i)}_{\text{perturbed}} - \mathrm{metric}^{(i)}_{\text{original}}
    \label{eq:metric_diff}
\end{equation}

Both perturbations induce directional hypotheses on score change $d_i$. For Persona Flip, responses are initially persona-aligned, so flipping constraints should reduce scores ($d_i<0$). For Sycophancy Attacks, replacing responses with performative variants likewise implies $d_i<0$. We report the mean/median of $d_i$, the decrease rate $\Pr(d_i<0)$, bootstrap 95\% CIs, and paired tests ($p$ values). To address within-case dependence, we repeat inference on case-level aggregates.

\subsection{Main Results}

\begin{figure}[htbp]
\raggedright
\captionsetup{justification=raggedright,singlelinecheck=false}
\captionsetup{justification=centering,singlelinecheck=true,font=small}
\caption{Sensitivity and robustness under controlled perturbations}
\raggedright
\footnotesize
\includegraphics[width=\linewidth]{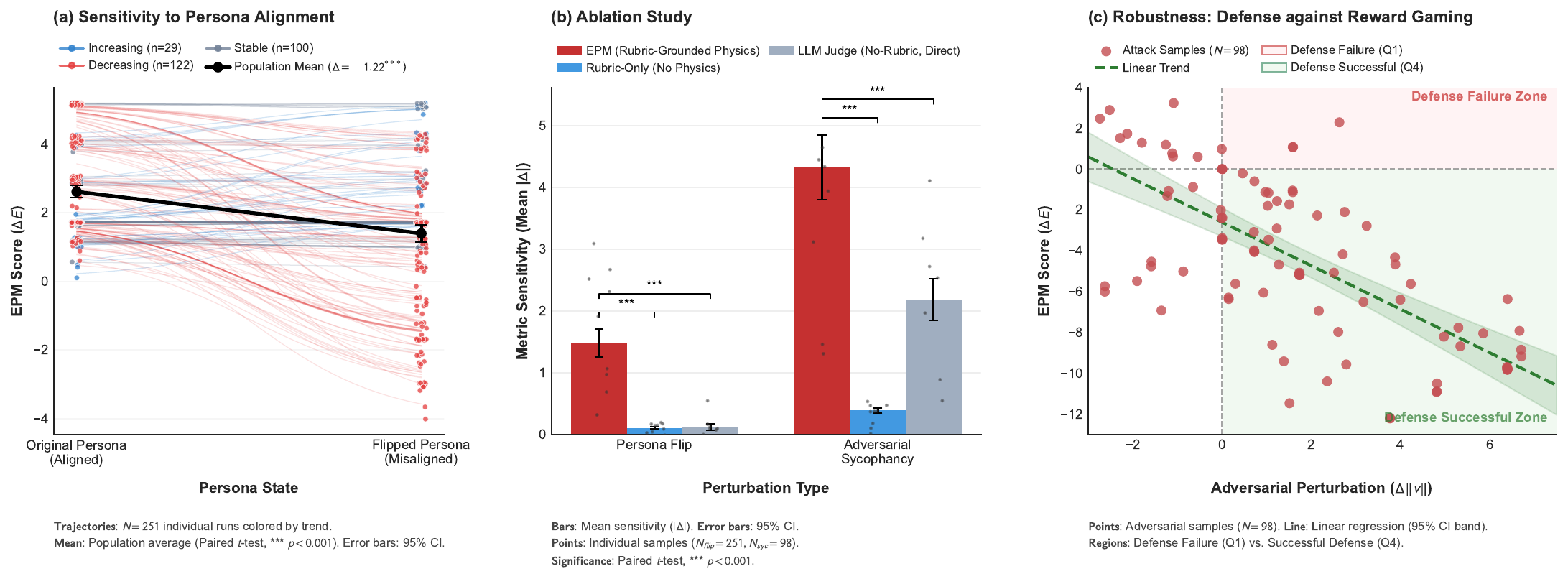}
\textbf{Note.} (a) Persona-conditional sensitivity: per-sample trajectories before and after flipping persona constraints (paired n = 251), showing a significant negative shift ($\Delta = -1.22$, \textit{p} < .0001).(b) Ablations: across Persona Flip and Sycophancy, Full EPM exhibits higher sensitivity, measured by mean absolute change, than a rubric-only variant and a direct LLM judge.(c) Adversarial robustness: under sycophancy, EPM penalizes replies that exhibit strong surface-level empathic signaling but fail to provide substantive support. The x-axis is perturbation magnitude ($\Delta \lVert v \rVert$) and the y-axis is the EPM score change.
\label{fig:controlled_perturbations}
\end{figure}

We first test persona-conditionality via Persona Flip and robustness to performative responses via Sycophancy Attacks. Results are summarized in Table~\ref{tab:main_results_full_epm} and visualized in Figure~\ref{fig:controlled_perturbations} (panels a and c).

\setcounter{table}{4}
\begin{table}[htbp]
\raggedright
\captionsetup{justification=centering,singlelinecheck=true,font=small}
\caption{Perturbation analysis ($\Delta$ = perturbed $-$ original)}
\label{tab:main_results_full_epm}
\small
\setlength{\tabcolsep}{4pt}
\renewcommand{\arraystretch}{1.15}
\begin{tabularx}{\linewidth}{*{8}{>{\centering\arraybackslash}X}}
\toprule
Dataset & $n$ & Metric & Mean & Median & $\Pr(d_i < 0)$ & 95\% CI & $p$ \\
\midrule
Persona Flip & 251 & $\Delta E$ & $-1.22$ & $-0.17$ & $72.50\%$ & $[-1.47,\,-0.97]$ & $< .0001$ \\
Persona Flip & 251 & $\Delta \cos\theta^{*}$ & $-0.38$ & $-0.03$ & $69.70\%$ & $[-0.47,\,-0.29]$ & $< .0001$ \\
Sycophancy & 98 & $\Delta E$ & $-4.36$ & $-4.68$ & $81.60\%$ & $[-5.15,\,-3.57]$ & $< .0001$ \\
Sycophancy & 98 & $\Delta \cos\theta^{*}$ & $-1.08$ & $-1.65$ & $84.70\%$ & $[-1.27,\,-0.90]$ & $< .0001$ \\
\bottomrule
\end{tabularx}

\vspace{4pt}
\begin{minipage}{0.95\linewidth}
\footnotesize
\noindent\textbf{Note.} Persona Flip is constructed from replies aligned under the original persona ($\Delta E > 0$ in $100.0\%$ of pairs), so the test targets whether counterfactually flipped constraints induce systematic deterioration. $\cos\theta^{*}$ denotes directional alignment between the intervention direction and the ideal healing direction.
\end{minipage}
\end{table}
\FloatBarrier

\textbf{Persona Flip: directional response to counterfactual priority inversion}

Persona Flip holds the response text fixed and counterfactually swaps persona-defined priorities and constraints. If EPM captures persona-conditioned support, replies aligned with the original persona should become misaligned after the flip, producing a one-sided score drop ($d_i<0$). A surface-based metric would be largely invariant to this change. On Persona Flip ($n = 251$), $\Delta E$ decreases with mean $-1.22$ and median $-0.17$, with a decrease rate of $72.5\%$ and a bootstrap $95\%$ confidence interval of $[-1.47,\,-0.97]$ ($p < .0001$). Alignment shows a consistent negative shift (mean $-0.38$; median $-0.03$; decrease rate $69.7\%$). Case-level aggregation yields the same conclusion (mean $-1.17$; $95\%$ confidence interval $[-1.86,\,-0.61]$; $p < .001$). These results support that EPM is persona-conditional even when the response text is held constant.

\textbf{Sycophancy Attacks: one-sided penalties for performative responses}

Sycophantic and templated replies often increase surface-level empathic signaling while contributing little to persona-consistent progress. Because our attacks are constructed as negative replacements, we expect a one-sided decrease. On Sycophancy Attacks ($n=98$), $\Delta E$ decreases sharply (mean $-4.36$; median $-4.68$; decrease rate $81.6\%$; 95\% CI $[-5.15,-3.57]$; $p<.0001$). Alignment also drops substantially (mean $-1.08$; median $-1.65$; decrease rate $84.7\%$). This indicates that EPM is not systematically fooled by surface-level empathic signaling.

\subsection{Ablation Study}

Main results establish directional sensitivity and robustness, but do not isolate which design elements are necessary. We therefore test two ablations: removing energy aggregation and replacing $\Delta E$ with single components (Figure~\ref{fig:controlled_perturbations}, panel b).

\textbf{Energy-aggregation ablation: reduced sensitivity and resolution}

We build a No-Physics variant that keeps rubric signals but replaces energy aggregation with a linear weighted score. If aggregation is essential, shifts should shrink and ties should increase; otherwise, paired shifts should remain similar. On Persona Flip, No-Physics yields a near-zero mean shift ($-0.09$; median $0.00$), a lower decrease rate ($46.2\%$), and a high tie rate ($44.2\%$), compared to $3.2\%$ ties for Full EPM. On Sycophancy, No-Physics shows only a small decrease (mean $-0.36$; median $-0.42$) with a $5.1\%$ tie rate. Overall, removing aggregation substantially reduces sensitivity and effective resolution.

\textbf{Component ablation: magnitude-only signals reward superficial intensity}

We further test whether $\Delta E$ can be replaced by a single factor. Under Persona Flip, the magnitude term shows minimal response and many ties ($52.2\%$). More critically, under sycophancy, the magnitude term increases (mean $1.65$; median $1.33$; only $25.5\%$ decreases), indicating that a strength-only signal systematically favors surface intensity, a failure mode EPM is designed to avoid.

\begin{table}[htbp]
\raggedright
\captionsetup{justification=centering,singlelinecheck=true,font=small}
\caption{Ablation results ($\Delta$ = perturbed $-$ original)}
\label{tab:ablation_effects}
\small
\setlength{\tabcolsep}{8pt}
\renewcommand{\arraystretch}{1.15}
\begin{tabularx}{\linewidth}{l c X r r c c}
\toprule
Dataset & $n$ & Component & Mean & Median & $\Pr(d_i < 0)$ & Tie rate \\
\midrule
Persona Flip & 251 & EPM (Rubric-Grounded Physics; $\Delta E$) & $-1.22$ & $-0.17$ & $72.50\%$ & $3.20\%$ \\
Persona Flip & 251 & Rubric-Only (No Physics) & $-0.09$ & $0.00$ & $46.20\%$ & $44.20\%$ \\
Persona Flip & 251 & LLM Judge (No-Rubric, Direct) & $-0.25$ & $0.00$ & $29.10\%$ & $52.20\%$ \\
\midrule
Sycophancy & 98 & EPM (Rubric-Grounded Physics; $\Delta E$) & $-4.36$ & $-4.68$ & $81.60\%$ & $3.10\%$ \\
Sycophancy & 98 & Rubric-Only (No Physics) & $-0.36$ & $-0.42$ & $85.90\%$ & $5.10\%$ \\
Sycophancy & 98 & LLM Judge (No-Rubric, Direct) & $1.65$ & $1.33$ & $25.50\%$ & $4.10\%$ \\
\bottomrule
\end{tabularx}

\vspace{4pt}
\begin{minipage}{0.95\linewidth}
\footnotesize
\noindent\textbf{Note.} $d_i$ is the paired change (perturbed $-$ original); $d_i<0$ means the perturbation is penalized. $\Pr(d_i<0)$ is the decrease rate, and \textit{Tie rate} counts near-zero changes. Full EPM penalizes both perturbations consistently with few ties. Rubric-Only largely collapses to near-zero shifts, while the direct/magnitude-only judge fails under sycophancy by increasing scores on average, reflecting a surface-intensity bias that EPM avoids.
\end{minipage}
\end{table}

\subsection{Summary}

Across paired controlled perturbations, EPM provides three lines of evidence: (i) persona-conditionality, demonstrated by systematic degradation under counterfactually flipped constraints with text held fixed; (ii) mechanistic necessity, shown by large sensitivity and resolution losses when removing energy aggregation; and (iii) robustness to performative responses, evidenced by strong one-sided penalties under sycophantic replacements. Human persona-proxy review is reported only as a supplementary pilot due to limited cross-persona generalizability (Appendix~\ref{sec:appendix_e}). Overall, this section validates the internal properties of EPM and its mechanism-level behavior.
\section{Results}
This section presents a comprehensive evaluation of 14 Large Language Models (LLMs) on the EMPA Benchmark. The multi-dimensional nature of the EPM framework allows us to move beyond aggregate scores and diagnose \textit{how} models succeed or fail within complex emotional dynamics. We analyze performance across four layers: (1) overall capability and stability, (2) mechanism adaptability (routine vs. challenging conditions), (3) persona resilience, and (4) process-level trajectory dynamics.

\subsection{Dataset}
The EPM Benchmark is designed as a stress test for empathetic dialogue systems rather than a standard capability probe. The full dataset contains over 1,000+ scenarios, each equipped with a complete Persona Card, memory archives, and plot background information. We use Multi-Dimensional Stratified Sampling to select 30 scenarios with orthogonal coverage across the C/A/P axes and six life domains. The set over-samples Hard/Extreme cases (86.7\%) and includes 50\% defensive personas, making the benchmark a stress test rather than a capability probe (see Figure~\ref{fig:dataset_overview} and Appendix~\ref{sec:appendix_a1}).

\subsection{Overall Performance}

We evaluated 14 models ranging from proprietary frontier systems (e.g., Claude 4.6 Opus, Gemini 3 Pro) to open-weights models (e.g., Llama 3.3, Qwen 3). Table~\ref{tab:epmq_ranking} presents the overall leaderboard based on the EPM-Q Score, a composite metric synthesizing Outcome Quality (40\%), Process Efficiency (20\%), and Stability (40\%).

\begin{table}[htbp]
\centering
\caption{EPM Benchmark Leaderboard}
\label{tab:epmq_ranking}
\resizebox{\textwidth}{!}{
\begin{tabular}{clcccccccccc}
\toprule
\multirow{2}{*}{Rank} & \multirow{2}{*}{Model} 
& \multicolumn{3}{c}{Outcome Quality} 
& \multicolumn{3}{c}{Process Efficiency} 
& \multicolumn{3}{c}{Strategic Stability} 
& \multirow{2}{*}{EPM-Q} \\
\cmidrule(lr){3-5} \cmidrule(lr){6-8} \cmidrule(lr){9-11}
&  & RDI & $E_{\mathrm{tot}}$ & $S_{\mathrm{net}}$ & $\rho$ & $S_{\mathrm{proj}}$ & $\tau$ & $R_{\mathrm{pos}}$ & Align & Pen &  \\
\midrule
1  & Claude 4.6 Opus         & 99.5 & 117.6 & 122.0 & 139.0 & 128.4 & 96.9 & 91.5 & 92.4 & 98.9 & \textbf{107.2} \\
2  & Gemini 3 Pro Preview    & 98.1 & 100.3 & 113.9 & 111.4 & 103.4 & 95.8 & 90.7 & 92.2 & 97.8 & \textbf{99.8} \\
3  & GPT-5.2 Pro             & 97.7 & 95.5  & 113.4 & 102.5 & 95.3  & 94.2 & 89.9 & 91.7 & 97.9 & \textbf{97.6} \\
4  & Gemini 2.5 Pro          & 92.6 & 90.1  & 124.2 & 65.5  & 62.2  & 81.4 & 88.2 & 87.1 & 93.8 & \textbf{90.7} \\
5  & Qwen 3 235B             & 92.0 & 91.7  & 121.3 & 88.3  & 82.6  & 74.0 & 80.2 & 82.0 & 82.3 & \textbf{89.6} \\
6  & Seed 2.0                & 90.9 & 87.2  & 134.2 & 55.7  & 53.1  & 70.8 & 83.1 & 84.0 & 88.1 & \textbf{87.6} \\
7  & Kimi k2                 & 87.3 & 86.6  & 123.2 & 72.6  & 68.3  & 70.2 & 80.1 & 81.6 & 82.1 & \textbf{86.2} \\
8  & Claude 3.5 Sonnet       & 91.9 & 83.3  & 128.3 & 54.8  & 52.2  & 72.3 & 79.1 & 81.5 & 84.7 & \textbf{85.1} \\
9  & DeepSeek V3             & 84.3 & 78.8  & 119.7 & 58.4  & 55.3  & 58.6 & 71.5 & 73.8 & 74.0 & \textbf{78.4} \\
10 & Seed 1.6                & 28.3 & 21.6  & 74.6  & 8.4   & 8.1   & 46.4 & 48.4 & 57.5 & 61.6 & \textbf{43.1} \\
11 & Llama 3.3 70B           & 27.0 & 10.0  & 76.0  & 5.2   & 5.0   & 30.1 & 48.3 & 55.5 & 52.2 & \textbf{38.5} \\
12 & GPT-4o                  & 18.7 & 14.3  & 61.4  & 5.7   & 5.5   & 58.4 & 38.2 & 47.0 & 38.7 & \textbf{33.7} \\
13 & Doubao 1.5              & 23.8 & 12.3  & 46.3  & 5.2   & 5.0   & 47.9 & 36.5 & 41.1 & 37.8 & \textbf{30.2} \\
14 & Qwen 3 32B              & 5.4  & 0.0   & 7.7   & 0.0   & 0.0   & 77.7 & 20.9 & 30.0 & 7.9  & \textbf{14.8} \\
15 & Llama 3.1 8B            & 2.6  & 0.0   & 0.4   & 0.0   & 0.0   & 82.6 & 19.5 & 27.7 & 15.8 & \textbf{14.3} \\
16 & Qwen 3 8B               & 1.2  & 0.0   & 0.9   & 0.0   & 0.0   & 85.0 & 16.1 & 25.5 & 5.0  & \textbf{12.2} \\
\bottomrule
\end{tabular}
}
\end{table}

The leaderboard reveals a clear four-tier stratification, where differences between tiers reflect distinct failure modes rather than simple capability gaps (see Section~\ref{sec:core_findings_four_laws}). A significant structural discontinuity exists between the top three models (Claude 4.6, Gemini 3, ChatGPT-5.2) and those ranked fourth through ninth, indicating a qualitative difference in underlying empathetic mechanisms.

\subsection{Core Findings: Four Laws of LLM Empathy}
\label{sec:core_findings_four_laws}

\textbf{Law 1. Outcome Efficiency Decoupling under Conservative Alignment}

The leaderboard data indicates that high outcome quality does not necessarily translate to a high EPM-Q composite score. Seed 2.0 achieves an outcome score (104.07) comparable to the second-ranked Gemini 3, yet ranks sixth due to a severely penalized efficiency score (59.84). This phenomenon reflects a common pattern in safety-aligned models, where the model defaults to low-risk soothing without making meaningful progress. Models tend to avoid substantial interventions to minimize failure risk, resulting in repetitive validation loops where the user's emotional entropy fails to converge effectively. In contrast, Claude 4.6 achieves the highest outcome quality with the highest efficiency (121.40), demonstrating that superior empathetic capability lies in knowing \textit{when} to transition from emotional pacing to proactive leading (see Appendix~\ref{sec:appendix_a2}).

\textbf{Law 2. Performance Stratification Driven by High Affective Entropy}

Mechanism stress tests reveal that performance on the Affective dimension (A-axis) under challenging conditions is the critical differentiator between model tiers. Top-tier models maintain high scores ($>100$) across both routine and challenging affective conditions. Mid-tier models (Seed 2.0, Kimi k2), however, show a significant drop (15–25 points) when shifting to challenging conditions. This degradation suggests that while mid-tier models have mastered standard empathy scripts, they lack the ability to dynamically recalibrate when user emotional entropy is extremely high. They tend to trigger the proactive axis (P) prematurely, before emotional resonance is fully established, thereby inducing user resistance. Furthermore, the unusually high scores of ChatGPT-5.2 Pro and Qwen 3 on Cognitive-Challenging (C-H) tasks may reflect an over-calibration effect in preference-tuned models, where responses become overly elaborate relative to the user’s underlying emotional needs (see Appendix~\ref{sec:appendix_a3}).

\textbf{Law 3. The Proactive Dimension Bottleneck in Complex Scenarios}

The Proactive-Challenging (P-H) condition proves to be the universal bottleneck for all evaluated models. Performance degradation in this condition is not limited to lower-tier models; even the fourth-ranked Gemini 2.5 Pro shows a marked decline, while models ranked ninth and below exhibit a precipitous drop. These results suggest that guiding a highly resistant user toward behavioral change requires not only emotional perception but also strategic timing and goal-oriented persuasion, capabilities that remain insufficiently developed in current training and alignment approaches (see Appendix~\ref{sec:appendix_a3}).

\textbf{Law 4. Generalization Challenges Posed by Defensive Personas}

Defensive users (characterized by high empathy thresholds and active psychological guarding) present a systematic challenge to all models, yet the \textit{magnitude} of this challenge defines model tiers. In the Affective-Defensive (A-Def) condition, only Claude 4.6 maintains a score above 105, while other models drop by more than 20 points. The failures of lower-tier models stem from performative empathy, namely standardized comforting phrases (e.g., "I understand how you feel") that defensive personas are designed to penalize. Notably, performance degradation is minimized in the Cognitive-Defensive condition, suggesting that defensive users may be more receptive to cognitive engagement paths. This offers a strategic insight for designing interventions for resistant users (see Appendix~\ref{sec:appendix_a4}).

\subsection{Model Stratification: Four Profiles of Empathetic Capability}

Based on our multi-dimensional analysis, the 14 models fall into four distinct profiles of empathetic capability:

\textbf{Tier 1: Precision Navigators}\\
These models (Claude 4.6 Opus, Gemini 3 Pro Preview, ChatGPT-5.2 Pro) demonstrate robust, precise, and efficient empathetic navigation, characterized by coherent trajectories that effectively employ pacing and leading strategies. In the 3D state space, they fully develop the Affective axis before decisively advancing along Cognitive and Proactive dimensions. Their radar charts exhibit a balanced, full hexagonal shape, reflecting high consistency across diverse domains and persona types. Despite their strength, they exhibit minor blind spots in the values \& beliefs domain and proactive-defensive conditions, with occasional inconsistency in extreme scenarios.

\textbf{Tier 2: Safe Stagnators} \\
While these models (Gemini 2.5 Pro, Qwen 3 235B, Seed 2.0, Kimi K2-0905, Claude 3.5 Sonnet, DeepSeek Chat V3) achieve outcome quality comparable to Tier 1, they are hampered by systematically low efficiency. Their trajectories often become trapped in prolonged validation loops, oscillating within the mid-range of the affective axis without effectively advancing into Cognitive or Proactive spaces. This safe but stagnant behavior is visually captured in their radar charts, which display large but irregular hexagons marked by distinct notches on the A-hard and P-hard axes.

\textbf{Tier 3: Capability Cliff }\\
A structural break separates these models (Seed 1.6, Llama 3.3 70B, ChatGPT-4o, Doubao 1.5 Character) from the upper tiers, with EPM-Q scores dropping precipitously to the 30–43 range. They exhibit a pattern of being locally effective yet globally unstable; their radar charts appear as severely atrophied triangles, indicating that competence is retained only in routine conditions while collapsing under challenging scenarios. Trajectory analysis reveals scattered and disordered paths with minimal success in complex emotional navigation.

\textbf{Tier 4: Systemic Failure} \\
With an EPM-Q score of 14.29 and near-zero outcome quality, this model (Llama 3.1 8B) represents a case of harmful failure. It is the only system evaluated that produces negative empathy effects, with trajectory analysis showing dialogue paths drifting away from the target origin and effectively worsening the user’s emotional state.The radar chart is almost entirely collapsed to the center, signaling a fundamental lack of capability across all domains, mechanisms, and personas.

\section{Conclusion and Future Work}
Our evaluation reveals a persistent gap between empathetic language and empathetic control. While modern LLMs often produce fluent, high-affect responses, they less reliably regulate interaction dynamics over time, especially under resistance and delayed, non-verifiable feedback. Across the four laws, a consistent implication emerges: empathetic intelligence is largely a scheduling problem, requiring latent-state tracking, timely intervention, and sustained directional commitment rather than isolated strong turns.

A key limitation of current training pipelines is that preference-optimized objectives can overweight short-term perceived helpfulness, favoring immediate comfort over long-horizon stabilization. This bias encourages safe but stagnant behaviors and weakens proactive intervention in high-resistance regimes. Future work should explore reward signals that capture process efficiency and trajectory-level progress, targeted data for defensive users and high-entropy scenarios, and alignment mechanisms such as persona alignment training to improve sustained, persona-consistent resonance. EMPA offers process- and trajectory-level signals to quantify these gains, enabling iterative and controlled optimization toward long-horizon performance.

\clearpage

\bibliographystyle{unsrt}
\bibliography{main,custom}

\clearpage
\beginappendix

\section{Detailed Chart Analysis}

\begin{figure}[h]
    \centering
    \includegraphics[width=\textwidth]{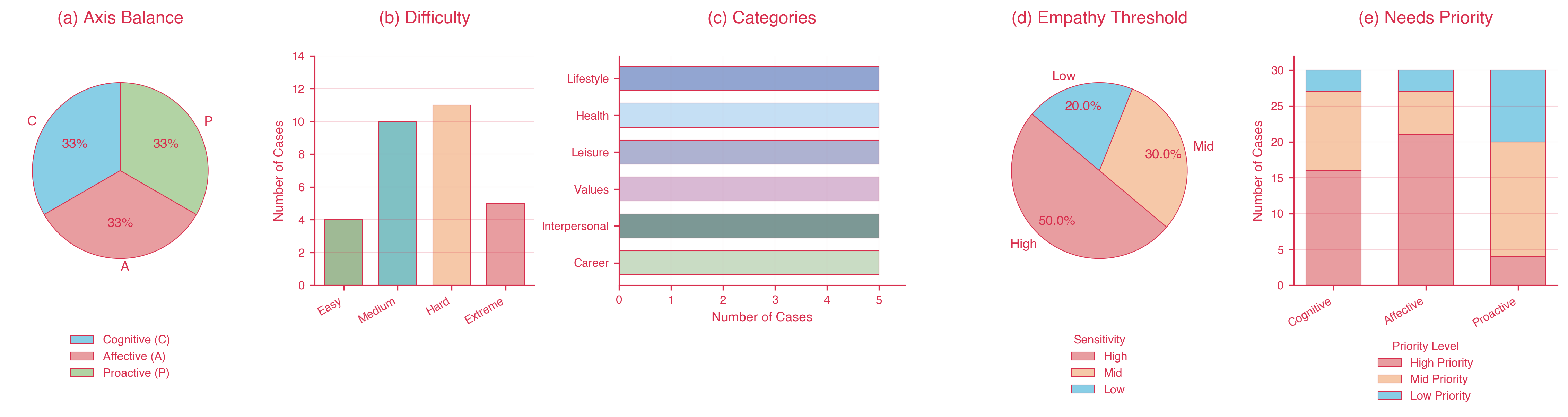}
    \caption{Structural distribution of the EPM Benchmark Dataset. (a) Perfect orthogonality across Cognitive (C), Affective (A), and Proactive (P) dimensions. (b) Right-skewed difficulty distribution, with 86.7\% of cases classified as Medium difficulty or above. (c) Coverage of six distinct life domains. (d-e) Demanding persona profile where 50\% of users possess a "High" empathy threshold.}
    \label{fig:dataset_overview}
\end{figure}

\subsection{Dataset Characteristics (Fig.~\ref{fig:dataset_overview})}
\label{sec:appendix_a1}

The dataset is constructed via multi-dimensional stratified sampling, ensuring balanced distribution: 10 cases each for Cognitive Restructuring (C-axis), Affective Resonance (A-axis), and Proactive Empowerment (P-axis), and 5 cases for each of the six life domains. Difficulty is quantitatively defined based on the initial empathy deficit ($||\vec{P}_0||$) distribution ($\mu=32.32, \sigma=4.52$): Extreme ($> \mu+\sigma$) 5 cases, Hard ($\mu$ to $\mu+\sigma$) 11 cases, Medium ($\mu-\sigma$ to $\mu$) 10 cases, Easy ($< \mu-\sigma$) 4 cases. The high proportion of Hard and Extreme scenarios ensures the benchmark's validity as a stress test. User persona analysis (Figure~\ref{fig:dataset_overview}d-e) further reveals that 50\% of simulated users hold a high empathy threshold, programmed to reject generic comfort or performative empathy—precisely targeting the failure modes of RLHF-aligned models.

\subsection{Detailed Analysis of Overall Performance and Stability}
\label{sec:appendix_a2}

\begin{figure}[h]
    \centering
    \includegraphics[width=\textwidth]{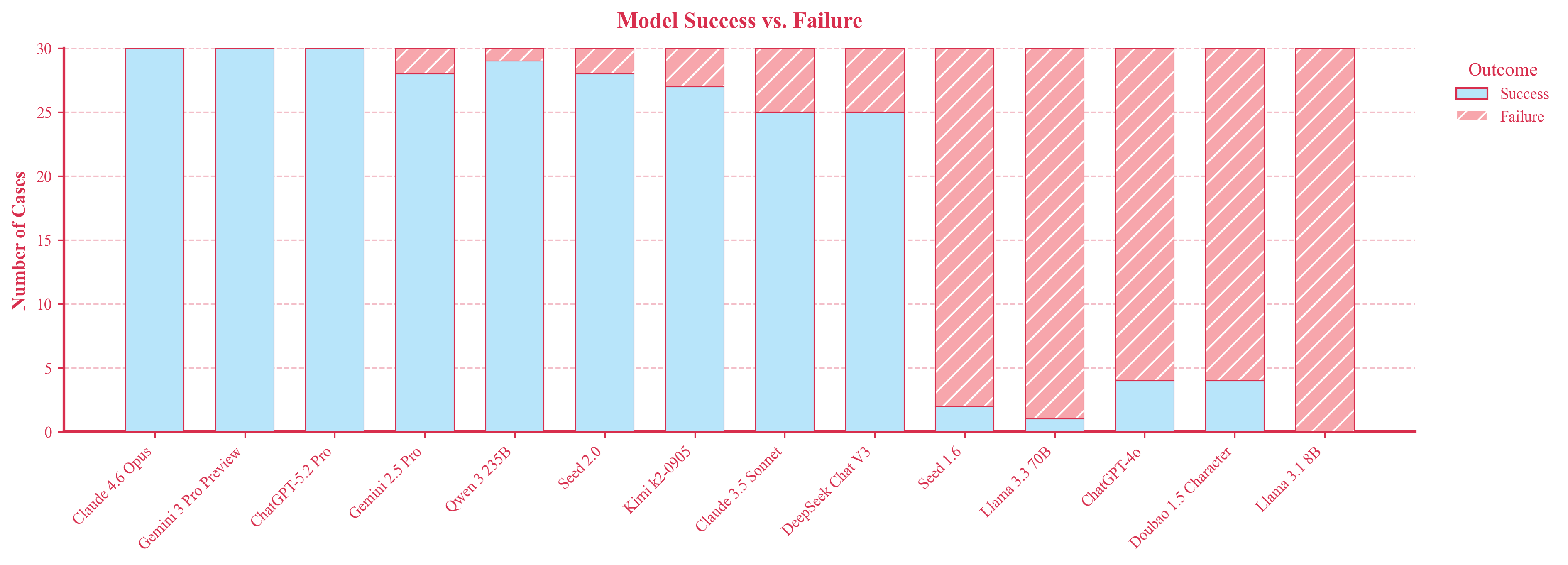}
    \caption{Stacked bar chart of Success (blue), Failure (pink), and Timeout (not shown) rates across models.}
    \label{fig:success_rate}
\end{figure}

\subsubsection{Success Rate Decomposition (Fig.~\ref{fig:success_rate})}

The decomposition of success rates distinguishes between two failure mechanisms: Explicit Failure (predominantly pink bars, e.g., Llama series), where models generate responses judged as harmful or emotionally misaligned, triggering explicit failure penalties; and Stagnation Failure (implied by lower success counts without explicit failure, e.g., Seed 2.0), where models rarely fail explicitly but frequently time out or exhaust turn limits without resolution. The prevalence of stagnation confirms the "Safe Stagnation" hypothesis in Section~\ref{sec:core_findings_four_laws}—models avoid failure by avoiding substantial action. While both mechanisms degrade user experience, their causes and remedies differ: explicit failure requires stricter safety alignment, while stagnation failure calls for "loosening" constraints to empower models to take calculated proactive risks.

\begin{figure}[h]
    \centering
    \includegraphics[width=\textwidth]{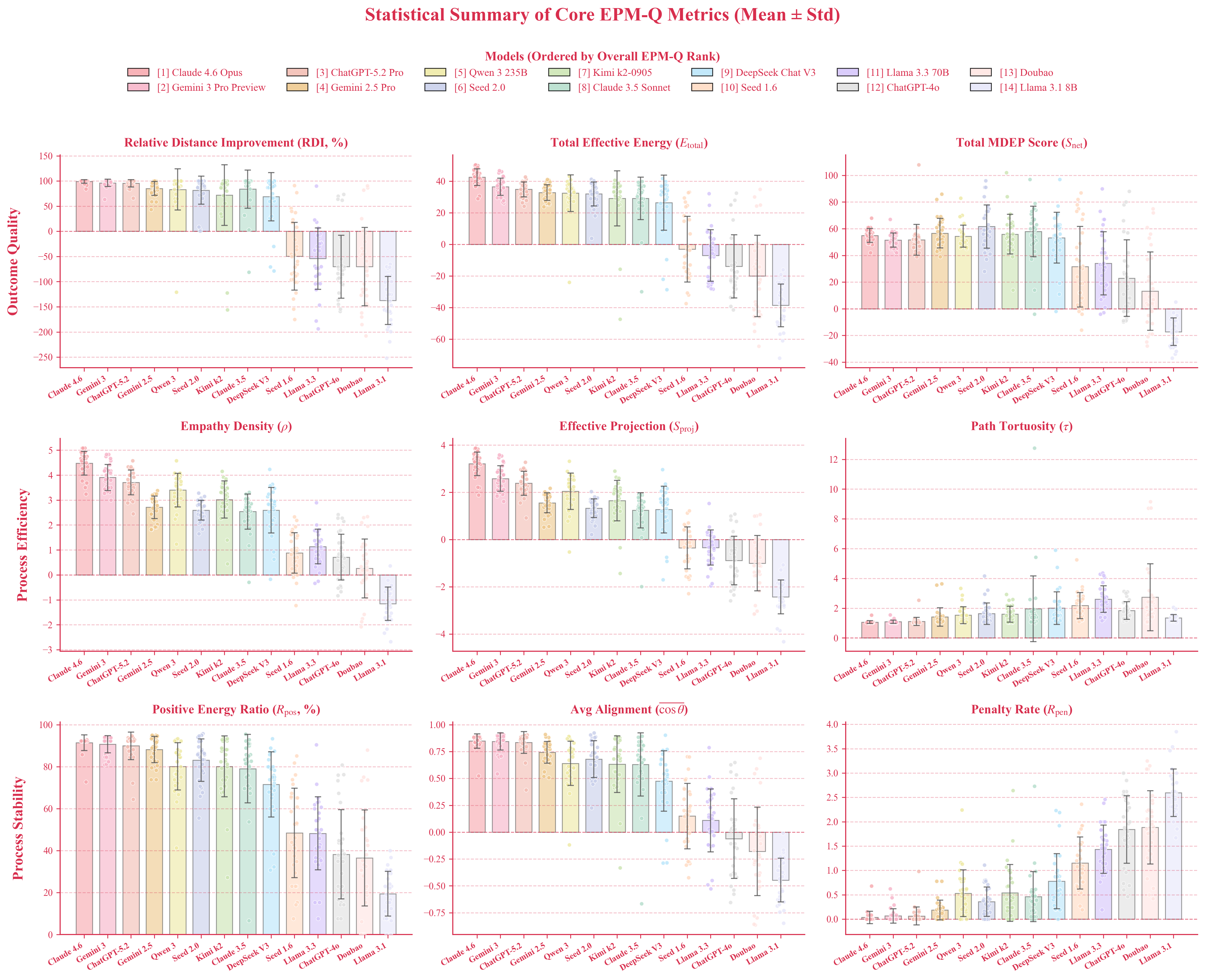}
    \caption{Statistical summary of core EPM-Q metrics (Mean $\pm$ Std). Panels display Outcome Quality (RDI, Total Effective Energy, Total MDEP Score), Process Efficiency (Empathy Density $\rho$, Effective Projection $S_{proj}$, Path Tortuosity $\tau$), and Process Stability (Positive Energy Ratio $R_{pos}$, Avg Alignment $\cos\theta$, Penalty Rate $R_{pen}$).}
    \label{fig:metrics_decomposition}
\end{figure}

\subsubsection{Decomposition of Nine Sub-Metrics (Fig.~\ref{fig:metrics_decomposition})}

The nine-panel visualization uncovers findings obscured by aggregate scores:

\begin{itemize}
    \item Relative Distance Improvement (RDI): Top-tier models achieve near 100\% RDI, whereas Llama 3.1 8B shows near-zero mean RDI with catastrophic variance, confirming systematic harm to user emotional states.
    \item Path Tortuosity ($\tau$): High tortuosity in Llama 3.3 70B and ChatGPT-4o indicates wandering rather than directed navigation in emotional space; Claude 4.6's lowest tortuosity confirms its therapeutic precision.
    \item Penalty Rate ($R_{pen}$): Disproportionately high penalties for Doubao 1.5 Character and Llama 3.1 8B are the single largest contributors to their EPM-Q collapse, reflecting frequent emotionally obtuse or rapport-breaking responses.
    \item Total Effective Energy ($E_{total}$) Variance: Kimi k2-0905's anomalously high variance reveals strategic instability—over-investing in some scenarios while under-investing in others.
\end{itemize}

\subsection{Detailed Performance Analysis}
\label{sec:appendix_a3}

\begin{figure}[h]
    \centering
    \includegraphics[width=\textwidth]{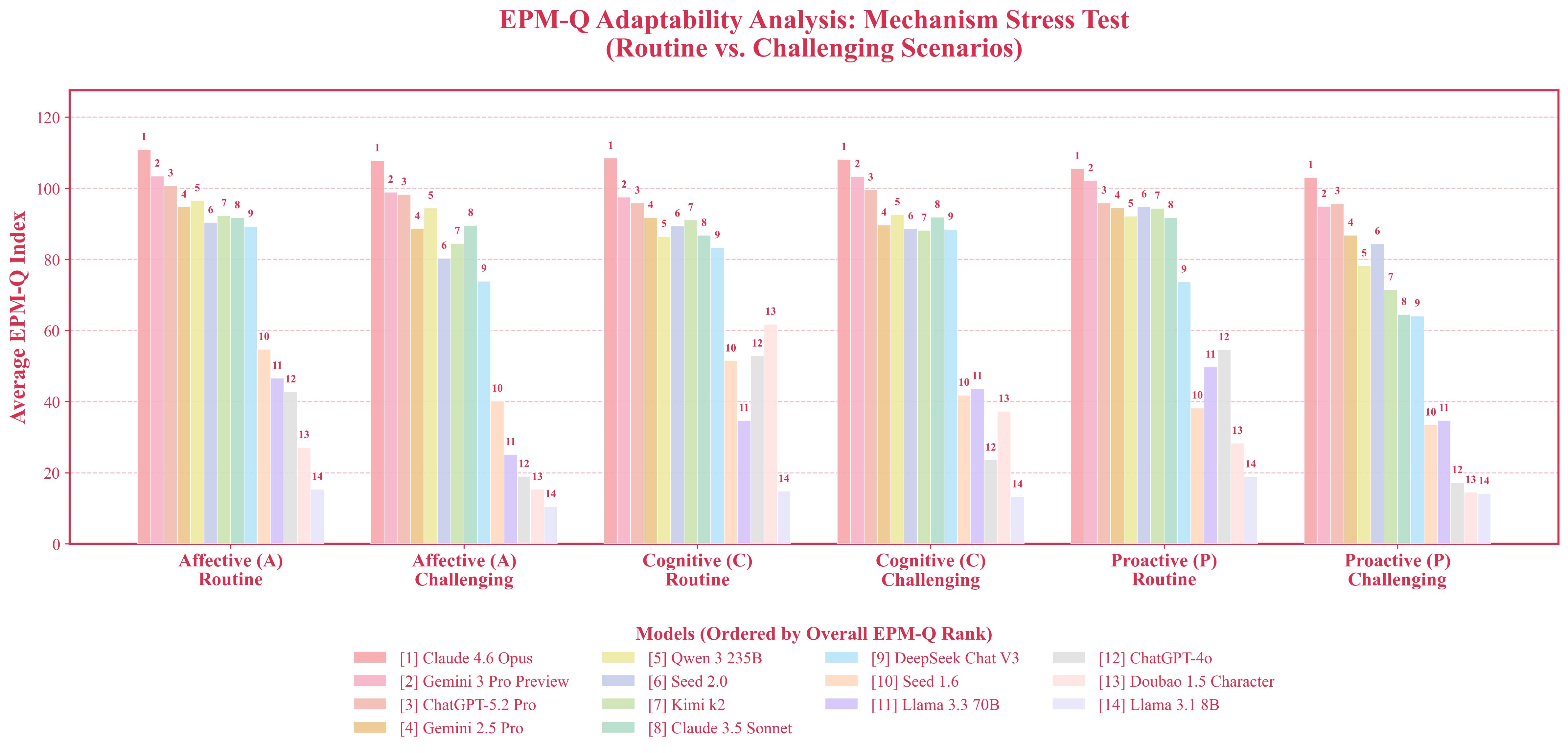}
    \caption{EPM-Q adaptability analysis. Comparison of model performance on Routine (light) vs. Challenging (dark) scenarios across Affective (A), Cognitive (C), and Proactive (P) axes.}
    \label{fig:mechanism_adaptability}
\end{figure}

\subsubsection{Mechanism Adaptability (Fig.~\ref{fig:mechanism_adaptability})}

Quantitative observations across conditions:
\begin{itemize}
    \item A-Challenging: Mid-tier models (Seed 2.0, Kimi k2) drop $\approx$15–25 points compared to A-Routine, while top-tier models maintain scores above 100.
    \item C-Challenging $>$ C-Routine: ChatGPT-5.2 Pro and Qwen 3 score higher on Cognitive-Challenging tasks than on Routine ones, supporting the RLHF Over-Calibration hypothesis—complex scenarios activate full capability, while simple ones trigger over-engineered responses.
    \item P-Challenging: This condition sees the sharpest decline across all models; Tier 2 models like Gemini 2.5 Pro show the most marked relative disadvantage here.
\end{itemize}

\begin{figure}[h]
    \centering
    \includegraphics[width=\textwidth]{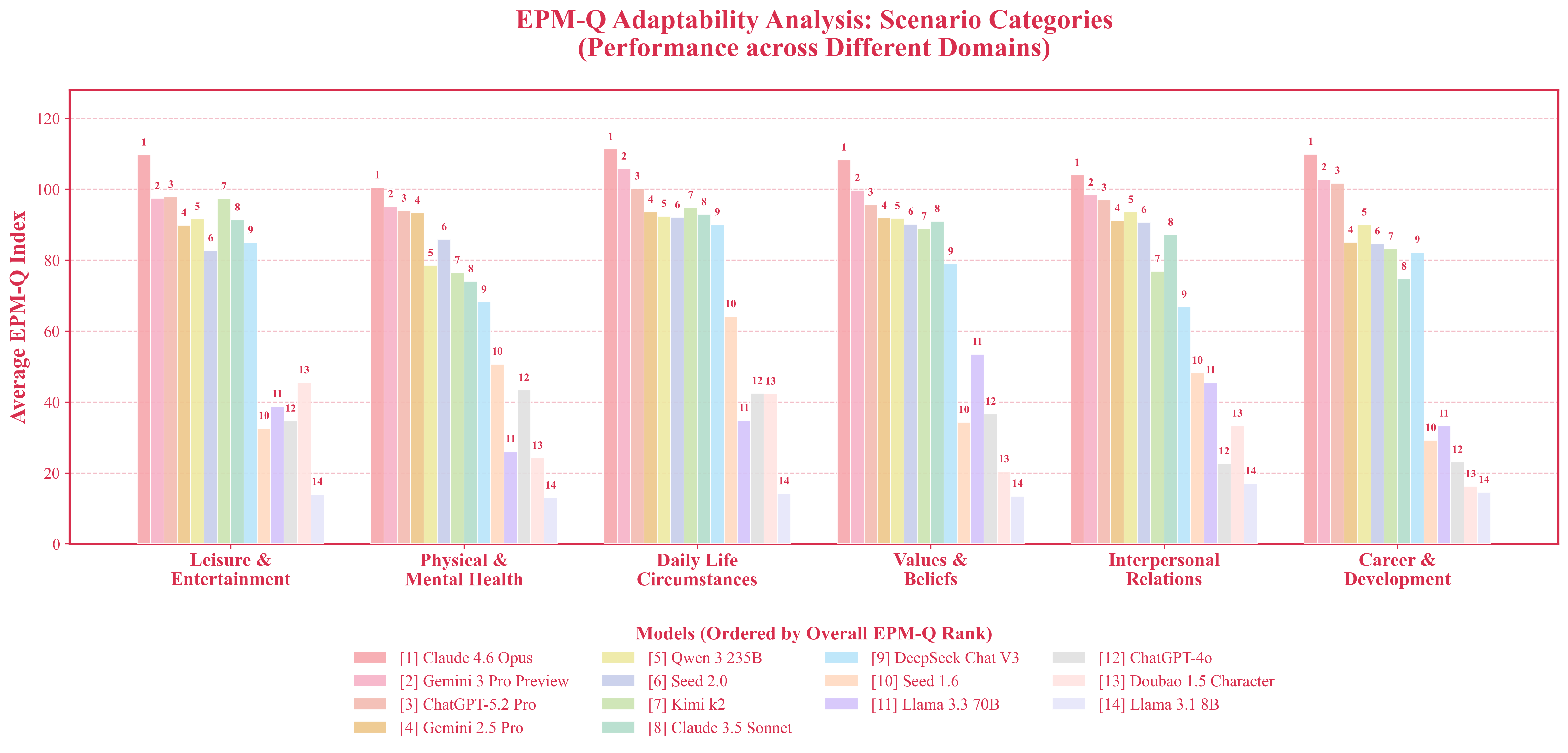}
    \caption{EPM-Q performance across six scenario domains.}
    \label{fig:scenario_category}
\end{figure}

\subsubsection{Scenario Category Analysis (Fig.~\ref{fig:scenario_category})}

Domain performance exhibits four recurring patterns. \textbf{Values \& Beliefs} shows the largest separation between top-tier and other models, with a bimodal score distribution, reflecting the need for ideological neutrality and precise reframing. \textbf{Physical \& Mental Health} shows the smallest gap among the top nine models, plausibly due to broader and more balanced pretraining coverage, although the bottom five models still degrade sharply. \textbf{Daily Life Circumstances} is most sensitive to efficiency, where models that default to low-risk, non-progressive responses perform worst. \textbf{Interpersonal Relations} most closely tracks the overall ranking and therefore serves as a reasonable proxy for general empathetic performance.

\begin{figure}[H]
    \centering
    \includegraphics[width=\textwidth]{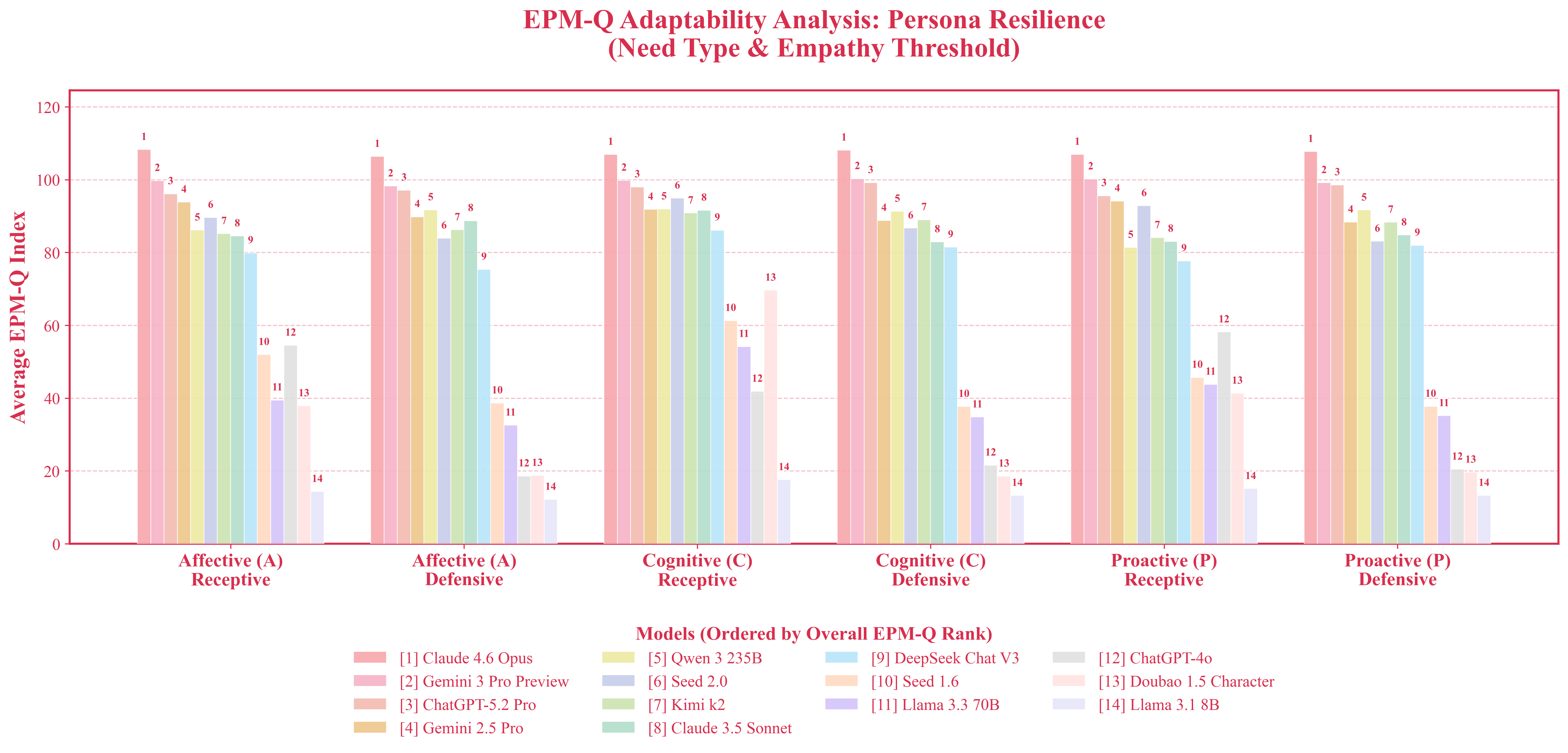}
    \caption{Performance breakdown by User Need Type (A/C/P) and Empathy Threshold (Receptive vs. Defensive).}
    \label{fig:persona_resilience}
\end{figure}

\FloatBarrier
\subsubsection{Persona Resilience Analysis (Fig.~\ref{fig:persona_resilience})}

Key quantitative findings: In the A-Def condition, Claude 4.6 maintains 105+ points, Gemini 3 and ChatGPT-5.2 maintain 95+, while all others fall below 90; Seed 2.0 and Kimi k2 drop over 20 points, marking their worst sub-category. The P-Def condition represents the absolute floor for all models, with the bottom five approaching zero or negative territory.

\begin{figure}[H]
    \centering
    \includegraphics[width=0.95\textwidth]{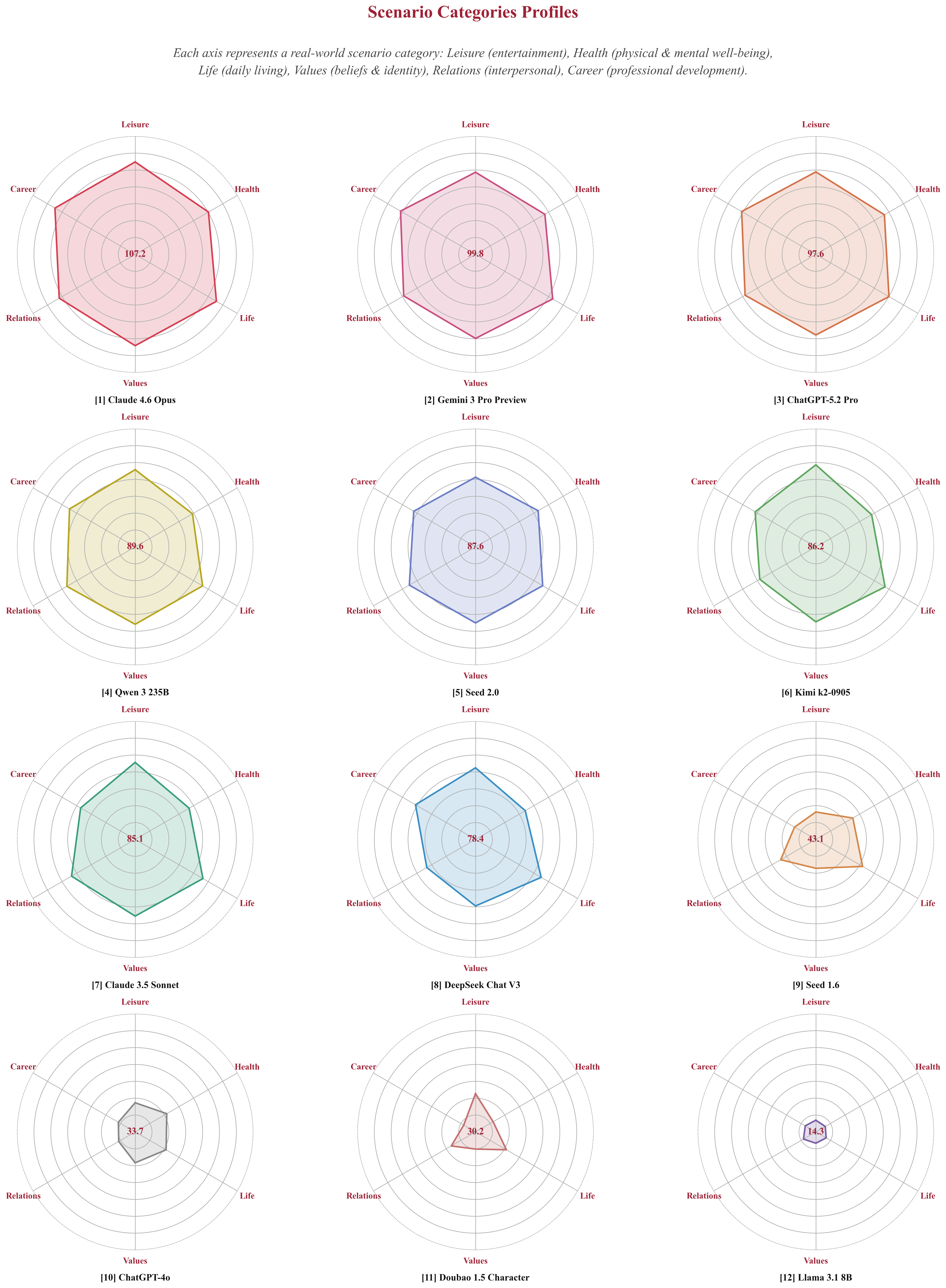}
    \caption{Radar chart grid of scenario category performance.}
    \label{fig:radar_categories}
\end{figure}

\begin{figure}[H]
    \centering
    \includegraphics[width=\textwidth]{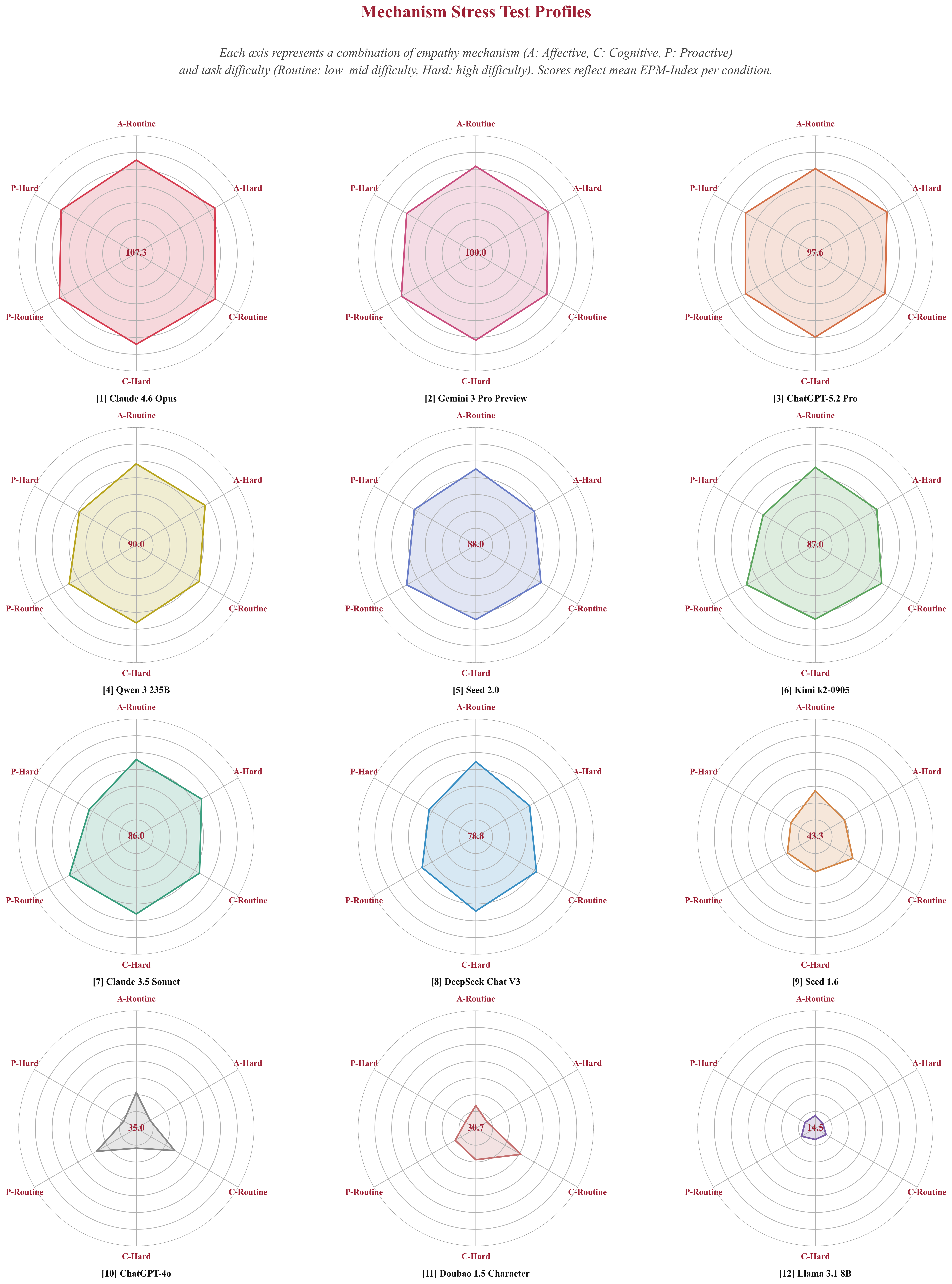}
    \caption{Radar chart grid of mechanism stress test profiles.}
    \label{fig:radar_mechanism}
\end{figure}

\begin{figure}[H]
    \centering
    \includegraphics[width=\textwidth]{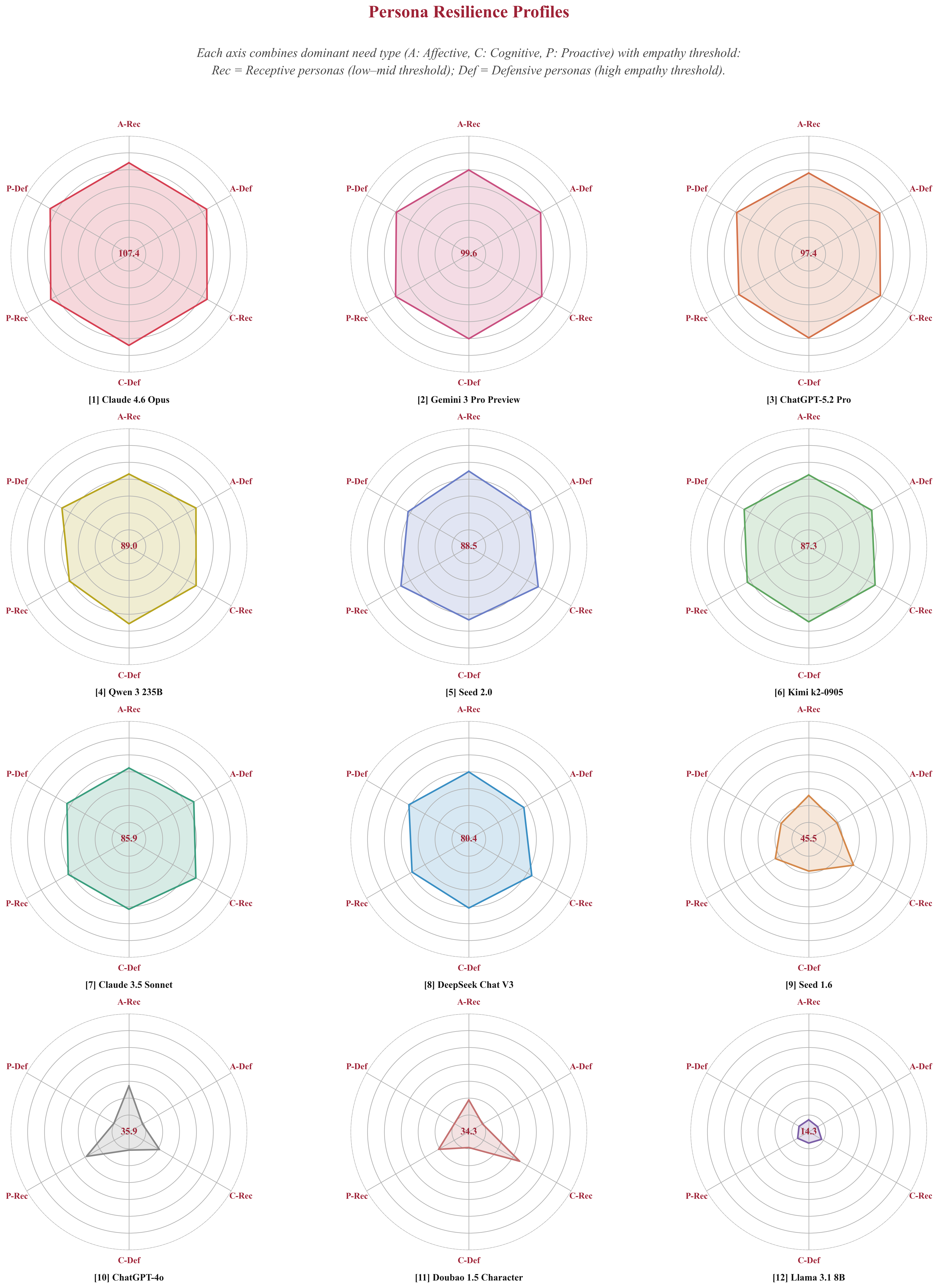}
    \caption{Radar chart grid of persona resilience profiles.}
    \label{fig:radar_persona}
\end{figure}

\FloatBarrier
\subsection{Holistic Profile Analysis (Radar Charts; Figs.~\ref{fig:radar_categories}--\ref{fig:radar_persona})}
\label{sec:appendix_a4}

\subsubsection{Scenario Category Radar Charts (Fig.~\ref{fig:radar_categories})}

Radar charts for scenario categories visually encode model versatility versus specialization. Claude 4.6’s near-perfect hexagon (radius 107.2) indicates domain-agnostic capability, maintaining high performance across all six axes. In contrast, Doubao 1.5 Character exhibits an extreme triangular profile with near-zero scores on Values and Relations axes, likely reflecting a training bias toward entertainment and light social scenarios. Llama 3.1 8B displays a globally atrophied polygon, confirming fundamental defects across all domains rather than localized weaknesses.

\subsubsection{Mechanism Radar Charts (Fig.~\ref{fig:radar_mechanism})}

Mechanism radar charts make the strategic profile of empathetic behavior explicit. Top-tier models form large, well-balanced polygons, indicating robust adaptability across both routine and challenging settings. Qwen 3 235B shows a pronounced notch on the A-Hard axis, consistent with an affective gap. Seed 2.0 exhibits a clear dip on the P-Hard axis, reflecting a tendency to default to low-risk, non-progressive responses. Lower-tier models display strong geometric asymmetries, often retaining only routine capabilities, or severely contracted polygons that indicate broad capability collapse across mechanisms.

\subsubsection{Persona Resilience Radar Charts (Fig.~\ref{fig:radar_persona})}
All models exhibit a consistent compression effect, with defensive axes systematically shorter than open axes. The extent of this compression is the primary differentiator across systems. Claude 4.6 retains an almost regular hexagon, providing a quantitative signature of strong persona resilience. Seed 1.6 collapses on the P-Def and A-Def axes while preserving positive C-Def scores, supporting a cognitive-first intervention strategy for defensive users. Doubao 1.5 Character shows an extreme single-axis profile, with performance concentrated almost entirely on A-Rec, representing the most severe persona brittleness observed in this evaluation.

\begin{figure}[H]
    \centering
    \includegraphics[width=\textwidth]{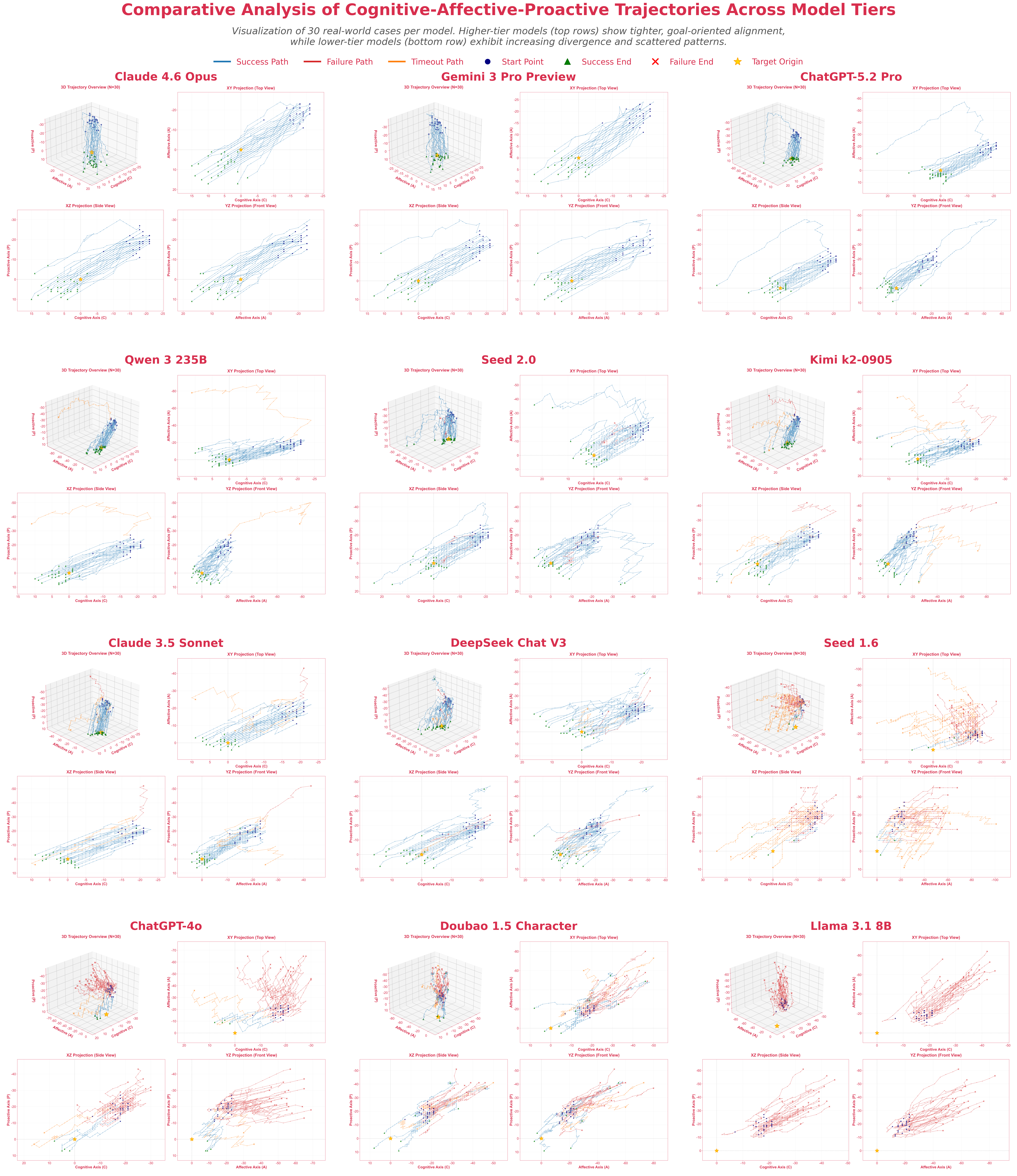}
    \caption{Comparative analysis of Cognitive-Affective-Proactive trajectories across model tiers.}
    \label{fig:trajectory_grid}
\end{figure}

\subsection{Process-Level Trajectory Analysis (Fig.~\ref{fig:trajectory_grid})}
\label{sec:appendix_a5}

This figure offers the richest diagnostic window, visualizing empathy strategies as geometric paths:

\begin{itemize}
    \item \textbf{Tier 1:} Trajectories form tightly bundled arcs that converge toward the target origin. In the XY top view, the dominant pattern is a pacing-and-leading arc, with an initial expansion followed by steady convergence. Failure traces are rare and concentrated near the target, consistent with near-miss errors rather than systematic collapse.
    \item \textbf{Tier 2:} Trajectory bundles remain coherent but exhibit prolonged oscillations around the mid-affective region, visible as dense horizontal bands in YZ side views. This pattern indicates validation loops. Success is typically reached, but along longer and less efficient paths.
    \item \textbf{Tier 3:} Successful trajectories preserve recognizable structure but occur less frequently, while dispersed failure paths become more common. In XZ side views, many failures show limited progress along the proactive axis, pointing to a specific deficit: affective regulation without effective proactive intervention.
    \item \textbf{Tier 4:} The 3D views are dominated by scattered, divergent trajectories with little shared structure. In the XY projection, Llama 3.1 8B shows paths that drift away from the target origin, making it the only model with negative mean empathy effects in this evaluation. Failure markers form a diffuse cloud far from the target, indicating broad loss of directional control.
\end{itemize}

\begin{figure}[H]
    \centering
    \includegraphics[width=\textwidth]{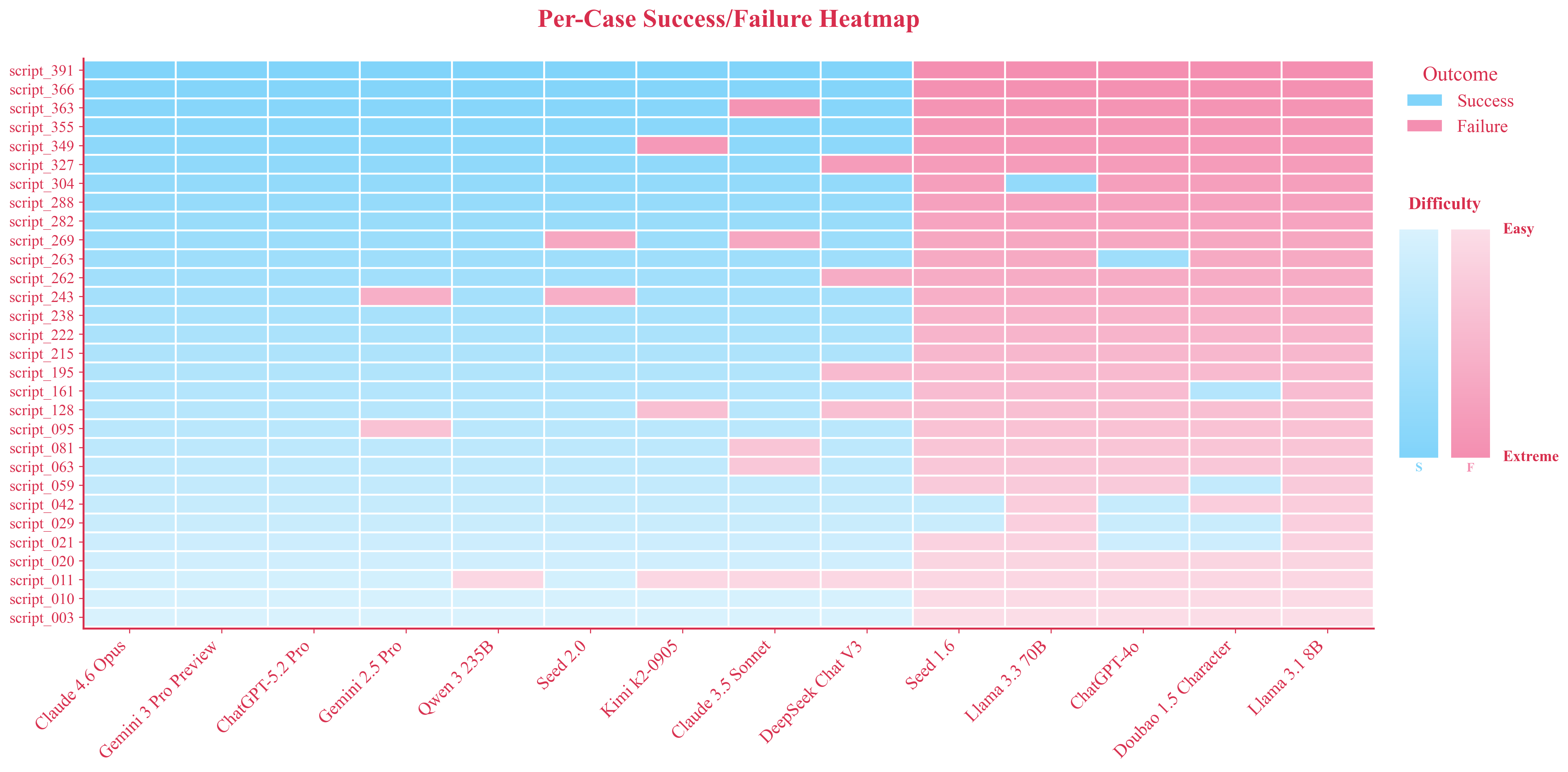}
    \caption{Heatmap of EPM-Q scores across all 30 test cases for all 14 models.}
    \label{fig:case_heatmap}
\end{figure}

\subsection{Granular Case Consistency (Fig.~\ref{fig:case_heatmap})}
\label{sec:appendix_a6}

The heatmap summarizes the consistency landscape across models and cases. Even top-tier systems exhibit localized weaknesses, with unexpected low-score regions concentrated in a small set of cases, most often within the Values \& Beliefs domain, aligning with the trend observed in the A.3 bar-chart analysis. Qwen 3 shows notable robustness on a subset of hard cases where other open-weight models degrade, consistent with the hypothesis that larger scale can provide a capacity buffer for inferring implicit psychological needs even when preference tuning is comparatively less polished than in proprietary systems. Finally, the global gradient from the top-left to the bottom-right of the matrix provides a clear visual signature of the four-tier stratification.

\section{EPM Metric Tables}
\label{sec:appendix_c}
\vspace{-6pt}
\begin{table}[H]
\centering
\small
\setlength{\tabcolsep}{6pt}
\renewcommand{\arraystretch}{1.35}
\caption{EPM Outcome Metrics -- Measuring Final Efficacy and Total Workload}
\label{tab:epm_outcome_metrics}
\begin{tabularx}{\linewidth}{l c X}
\toprule
Metric Name & Symbol & Core Meaning and Evaluation Value \\
\midrule
Task Completion Status & $\mathrm{Status}$ & Success/failure is defined by the Trinity Victory Condition—meeting the geometric/positional goal and accumulating sufficient energy. \\
Relative Distance Improvement & $\mathrm{RDI}$ & Measures the thoroughness of healing. Calculates the percentage improvement of the user's final psychological state relative to the initial deficit. \\
Cumulative Effective Energy & $E_{\text{total}}$ & Measures cumulative effective intervention along the ideal healing direction across the dialogue, serving as a proxy for substantive effort. \\
Energy Surplus & $E_{\text{surplus}}$ & Measures empathy abundance. Calculates the additional energy support provided beyond the basic requirement. \\
Total MDEP Net Score & $S_{\text{net}}$ & Measures total empathy quality. The sum of cumulative net scores obtained in the three dimensions of C/A/P. \\
\bottomrule
\end{tabularx}
\end{table}

\vspace{-6pt}
\begin{table}[H]
\centering
\small
\setlength{\tabcolsep}{6pt}
\renewcommand{\arraystretch}{1.35}

\caption{EPM Process Metrics}
\label{tab:epm_process_metrics}

\begin{subtable}{\linewidth}
\caption{EPM Process Efficiency Metrics -- Measuring Time Cost and Strategic Directness}
\label{tab:epm_process_efficiency_metrics_a}
\centering
\begin{tabularx}{\linewidth}{l c X}
\toprule
Metric Name & Symbol & Core Meaning and Evaluation Value \\
\midrule
Empathy Density & $\rho$ & Measures average intervention intensity. The ``gold content'' of effective empathy energy delivered on average per dialogue turn. \\
Average Effective Projection & $S_{\text{proj}}$ & Measures single-turn effectiveness. The average effective projection component of the action vector along the ideal direction per turn, serving as the basic unit of effective energy. \\
Path Tortuosity & $\tau$ & Measures strategic directness. The ratio of the actual action trajectory length to the straight-line displacement between start and end points, reflecting whether the strategy is efficient direct access or circuitous trial-and-error. \\
\bottomrule
\end{tabularx}
\end{subtable}

\vspace{6pt} 

\begin{subtable}{\linewidth}
\caption{EPM Process Stability Metrics -- Measuring Interaction Smoothness, Directional Correctness, and Safety}
\label{tab:epm_process_stability_metrics_b}
\centering
\begin{tabularx}{\linewidth}{l c X}
\toprule
Metric Name & Symbol & Core Meaning and Evaluation Value \\
\midrule
Average Alignment & $\overline{\cos\theta}$ & Measures directional consistency. The average cosine value of the angle between the model's intervention direction and the ideal healing direction. \\
Positive Energy Ratio & $R_{\text{pos}}$ & Measures process smoothness. The proportion of turns generating positive propulsion out of total turns. \\
Performative Penalty Rate & $R_{\text{pen}}$ & Measures the intensity of negative behavior. Quantifies the average intensity of punishment received by the model per turn due to inappropriate remarks (e.g., lecturing, indifference). \\
\bottomrule
\end{tabularx}
\end{subtable}

\vspace{4pt}
\begin{minipage}{0.95\linewidth}
\footnotesize
\noindent\textbf{Note.} We adopt an open, comprehensive quantitative evaluation paradigm. Metric weights are scenario-dependent and can be adjusted for capability profiling, instead of relying on a single aggregate ranking.
\end{minipage}
\end{table}
\vspace{-6pt}
\FloatBarrier
\vspace{12pt}
\section{EPM-Q Calculation Details and Mathematical Definitions}
\label{sec:appendix_b}

The EPM-Q (Empathy Physics Model - Quantitative Score) system transforms raw psychodynamic vectors into a standardized metric via a case-by-case normalization paradigm. This appendix provides the formal definitions, symbol explanations, and aggregation protocols.

\subsection{Fundamental Scientific Constants \& Derivations}

The calculation relies on three physically defined anchors to ensure scale invariance across diverse scenarios.

\textbf{1. Case-Specific Physical Benchmark ($r_{0,i}$).}

For any test case $i$, the difficulty is strictly defined by the $\ell_2$-norm of the user's initial psychological state vector $P_{0,i}$:
\begin{equation}
r_{0,i}=\lVert P_{0,i}\rVert_2
=\sqrt{c_{0,i}^2+a_{0,i}^2+p_{0,i}^2}.
\end{equation}
Where:
\begin{itemize}
  \item $P_{0,i}=[c_{0,i},\,a_{0,i},\,p_{0,i}]^\top$ denotes the initial state vector for case $i$;
  \item $c_{0,i}$, $a_{0,i}$, and $p_{0,i}$ denote the initial deficits in the Cognitive, Affective, and Proactive dimensions, respectively.
  \item \textbf{Remark:} For numerical stability, we add a small constant $\epsilon=10^{-6}$ to the denominator in any division involving $r_{0,i}$.
\end{itemize}

\textbf{2. Global Theoretical Maximum Intensity ($\rho_{\max}$).}

Defined as the theoretical ceiling of instantaneous empathetic power within the MDEP scale boundaries ($[-2, +2]$ per axis):
\begin{equation}
\rho_{\max}=\sup_{\vec{v}\in\mathcal{V}}\lVert\vec{v}\rVert_2
=\sqrt{2^2+2^2+2^2}\approx 3.464.
\end{equation}
Where:
\begin{itemize}
  \item $\mathcal{V}=[-2,2]^3$ is the bounded action space defined by the MDEP rubric;
  \item $\vec{v}$ represents an arbitrary single-turn action vector.
\end{itemize}

\textbf{3. Physical-to-Scale Conversion Factor ($\alpha$).}

The factor $\alpha$ calibrates the relationship between scalar score summation ($\ell_1$-like norm) and vector displacement ($\ell_2$ norm). In a 3D Euclidean space, the relationship between norms is bounded by the Cauchy--Schwarz inequalities:
\begin{equation}
\lVert\vec{x}\rVert_2 \le \lVert\vec{x}\rVert_1 \le \sqrt{3}\cdot \lVert\vec{x}\rVert_2.
\end{equation}
Where:
\begin{itemize}
  \item $\lVert\vec{x}\rVert_1$ denotes the Manhattan norm (sum of absolute components);
  \item $\lVert\vec{x}\rVert_2$ denotes the Euclidean norm.
  \item We set $\alpha\approx 1.2$ as the calibrated constant for therapeutic trajectories, representing the realistic distribution of strategic focus between single-axis intervention and holistic support.

\end{itemize}

\subsection{Normalization Formulas (Case-Level)}

\subsubsection{Unbounded Cumulative Metrics (Outcome Quality)}

These metrics measure the total work performed relative to the specific case difficulty $r_{0,i}$.

\textbf{1. Cumulative Energy Index ($Idx_{E_{\mathrm{tot}}}$).}

Defined as the cumulative effective energy performed (clipped at zero) normalized by the case-specific difficulty $r_{0,i}$:
\begin{equation}
Idx_{E_{\mathrm{tot}}, i}
=\frac{\max\!\left(0,\,E_{\mathrm{total},i}\right)}{r_{0,i}}\times 100.
\end{equation}
Where:
\begin{itemize}
  \item $E_{\mathrm{total},i}=\sum_{t=1}^{T_i}\Delta E_{t,i}$ is the cumulative effective energy accumulated over $T_i$ turns.

  \item \textbf{Note:} This index is intentionally unbounded to capture \emph{excellence}(performance exceeding the minimum deficit requirements).

\end{itemize}

\textbf{2. Total Net Score Index ($Idx_{S_{\mathrm{net}}}$).}

Defined as the total net scalar score (clipped at zero) normalized by the calibrated physical-to-scale factor $\alpha$ and the case-specific difficulty $r_{0,i}$:
\begin{equation}
Idx_{S_{\mathrm{net}}, i}
=\frac{\max\!\left(0,\,S_{\mathrm{net},i}\right)}{\alpha\cdot r_{0,i}}\times 100.
\end{equation}
Where:
\begin{itemize}
  \item $S_{\mathrm{net},i}=\sum_{t=1}^{T_i}\sum_{j\in\{C,A,P\}} v_{t,j}$ is the scalar sum of net scores across all dimensions.
\end{itemize}

\subsubsection{Unbounded Intensity Metrics (Process Efficiency)}

These metrics normalize interaction intensity against the theoretical limit $\rho_{\max}$.

\textbf{1. Empathy Density Index ($Idx_{\rho}$).}

Defined as the average effective energy per turn (clipped at zero) normalized by the global theoretical maximum intensity $\rho_{\max}$:
\begin{equation}
Idx_{\rho,i}
=\frac{\max\!\left(0,\,\rho_i\right)}{\rho_{\max}}\times 100.
\end{equation}
Where:
\begin{itemize}
  \item $\rho_i=E_{\mathrm{total},i}/T_i$ represents the average effective energy per turn.

  \item \textbf{Regularization Proof:} This metric inherently regularizes the unbounded $Idx_{E_{\mathrm{tot}}}$. Since $\rho_i$ is inversely proportional to $T_i$, any attempt to artificially inflate cumulative energy by extending the conversation length $T_i$ without maintaining high-quality intensity will incur a proportional penalty in $Idx_{\rho}$.
\end{itemize}

\textbf{2. Effective Projection Index ($Idx_{S_{\mathrm{proj}}}$).}

Defined as the average projection score onto the ideal gradient (clipped at zero) normalized by the global theoretical maximum intensity $\rho_{\max}$:
\begin{equation}
Idx_{S_{\mathrm{proj}}, i}
=\frac{\max\!\left(0,\,S_{\mathrm{proj},i}\right)}{\rho_{\max}}\times 100.
\end{equation}
Where:
\begin{itemize}
  \item $S_{\mathrm{proj},i}$ is the average projection of action vectors onto the ideal gradient.
\end{itemize}

\subsubsection{Bounded Ratio Metrics (Stability \& Strategy)}

These metrics are mapped to a standardized $[0,100]$ scale using a unified linear interpolation function.

\textbf{1. General Mapping Formula.}

Let $x$ be the raw metric value. We define the standardization function $\Phi$ that maps a raw range to the index score $[0,100]$:
\begin{equation}
\Phi(x;\,x_{0},x_{100})
=\mathrm{Clamp}\!\left(
\frac{x-x_{0}}{x_{100}-x_{0}}\times 100,\,
0,\,
100
\right).
\end{equation}
Where:
\begin{itemize}
  \item $x_{0}$ represents the physical boundary corresponding to a score of $0$ (Worst Case);
  \item $x_{100}$ represents the physical boundary corresponding to a score of $100$ (Best Case).
  \item This formulation automatically handles both forward metrics ($x_{100}>x_{0}$) and reverse metrics ($x_{100}<x_{0}$).
\end{itemize}

\textbf{2. Metric Specifications.}

The specific boundaries for each metric are defined as follows:

\begin{table}[H]
\centering
\small
\setlength{\tabcolsep}{6pt}
\renewcommand{\arraystretch}{1.25}
\caption{Metric Specifications for Unified Linear Mapping}
\label{tab:metric_specifications}
\begin{tabular}{p{4.0cm} c c p{4.0cm} p{4.0cm}}
\toprule
Metric Name & Symbol & Raw Range & Boundary $x_{0}$ (Score 0) & Boundary $x_{100}$ (Score 100) \\
\midrule
Relative Dist. Improvement & $Idx_{\{RDI\}}$   & $[-1,1]$ & $-1.0$ (Deterioration)       & $1.0$ (Full Resolution) \\
Alignment Index             & $Idx_{\{Align\}}$ & $[-1,1]$ & $-1.0$ (Opposite)            & $1.0$ (Perfect Alignment) \\
Path Tortuosity             & $Idx_{\{\tau\}}$   & $[1,3]$  & $3.0$ (Inefficient)          & $1.0$ (Optimal) \\
Penalty Rate                & $Idx_{\{Pen\}}$    & $[0,3]$  & $3.0$ (High Toxicity)        & $0.0$ (Zero Penalty) \\
\bottomrule
\end{tabular}
\end{table}

\subsection{Aggregation \& Final Score}

\subsubsection{Dimension Synthesis}

Let $N$ be the total number of test cases. Dataset-level averages ($\tilde{S}$) are grouped into three core dimensions:
\begin{equation*}
\mathcal{S}_{Outcome}=\frac{1}{3}\left(\tilde{S}_{RDI}+\tilde{S}_{E_{\mathrm{tot}}}+\tilde{S}_{S_{\mathrm{net}}}\right)
\end{equation*}
\begin{equation*}
\mathcal{S}_{Efficiency}=\frac{1}{3}\left(\tilde{S}_{\rho}+\tilde{S}_{S_{\mathrm{proj}}}+\tilde{S}_{\tau}\right)
\end{equation*}
\begin{equation*}
\mathcal{S}_{Stability}=\frac{1}{3}\left(\tilde{S}_{R_{\mathrm{pos}}}+\tilde{S}_{Align}+\tilde{S}_{Pen}\right)
\end{equation*}
Where:
\begin{equation}
\tilde{S}_{\mathrm{Metric}}=\frac{1}{N}\sum_{i=1}^{N} Idx_{\mathrm{Metric},i}.
\end{equation}

\subsubsection{Final EPM-Q Score}

To ensure scientific rigor in scoring, the aforementioned raw physical metrics are not directly summed up but converted following a set of Scientifically-Defined Open Benchmark Index logic.

\textbf{1. Scientific Anchoring.} All calculation benchmarks are strictly anchored to the physical definition of the task (such as initial deficit $r_{0}$) or the mathematical theoretical limit of the scale (such as maximum intensity $\rho_{\max}$), rather than arbitrary empirical values.

\textbf{2. Classification Conversion.}
\begin{itemize}
  \item For unbounded cumulative metrics (such as energy and net score), their multiplier relative to the scientific benchmark is calculated, forming an uncapped open index to reflect the excess performance of exceptional models.
  \item For bounded ratio metrics (such as RDI and alignment), standard linear mapping is adopted to convert their physical boundaries into $[0,100]$ interval scores.
\end{itemize}

\textbf{3. Synthetic EPM-Index.} Finally, through weighted synthesis, an open EPM benchmark index is output. $\mathrm{Index}=100$ represents precisely achieving the scientific benchmark, while $\mathrm{Index}>100$ intuitively reflects the excellence multiplier beyond the benchmark:
\begin{equation}
\mathbf{EPM\text{-}Index}
=0.4\cdot\mathbf{\tilde{S}_{Outcome}}
+0.2\cdot\mathbf{\tilde{S}_{Efficiency}}
+0.4\cdot\mathbf{\tilde{S}_{Stability}}.
\end{equation}

\vspace{12pt}
\section{Persona Card}
\label{sec:appendix_d}

\tcbset{colback=nsblue!10!white, colframe=nsblue, width=\linewidth, arc=5mm, breakable}
\begin{tcolorbox}[breakable]
\subsection*{Persona Card Schema}
\vspace{-5pt}

\textbf{Persona Card: }

\textbf{Role Information}
\begin{itemize}
  \item {Name:} 
  \item {Gender:} 
  \item {Age:} 
\end{itemize}

\textbf{Role Traits}
\begin{itemize}
  \item {Social persona:} 
  \item {Inner core:} 
\end{itemize}

\textbf{Baseline Empathy Threshold}

\textbf{Empathy threshold: [Medium].}

\textbf{Chat Topic}

\textbf{Empathy Needs}
\begin{itemize}
  \item {What she wants to vent:} 
  \item {Empathic points she hopes to receive:} 
  \item {Empathy threshold constraints:} 
\end{itemize}

\textbf{Current Empathy Priority}
\begin{itemize}
  \item {Affective empathy: [Priority: ].} 
  \item {Motivational empathy: [Priority: ].} 
  \item {Cognitive empathy: [Priority: ].} 
\end{itemize}

\textbf{Past Experiences}
\begin{itemize}
  \item {Childhood:} 
  \item {Adolescence:} 
  \item {Young adulthood:} 
  \item {Implicit growth arc:} 
\end{itemize}

\textbf{Current Situation}
\begin{itemize}
  \item {Present circumstances:} 
  \item {Main life goal at present:} 
  \item {Vision and motivation:} 
\end{itemize}

\textbf{Story}

\textbf{Trigger}

\textbf{Development}
\begin{itemize}
  \item {Stage 1: Evoked memory.} 
  \item {Stage 2: Reflection.} 
  \item {Stage 3: Self-examination.} 
  \item {Stage 4: Emotional eruption.} 
\end{itemize}

\textbf{Outcome}

\textbf{Epilogue}
\end{tcolorbox}

\tcbset{colback=nsblue!10!white, colframe=nsblue, width=\linewidth, arc=5mm, breakable}
\begin{tcolorbox}[breakable]
\subsection*{Persona Card Example}
\vspace{-5pt}

\textbf{Persona Card: Lin Xiaoyue}

\textbf{Role Information}
\begin{itemize}
  \item \textbf{Name:} Lin Xiaoyue
  \item \textbf{Gender:} Female
  \item \textbf{Age:} 23
\end{itemize}

\textbf{Role Traits}
\begin{itemize}
  \item \textbf{Social persona:} Outgoing and independent. She enjoys sharing study progress with friends, but rarely reveals vulnerability or insecurity. She is accustomed to presenting herself as resilient and well-planned.
  \item \textbf{Inner core:} Strongly goal-oriented with a high drive for self-improvement. Deep down, she fears failure and is prone to anxiety and self-doubt when she cannot see immediate returns.
\end{itemize}

\textbf{Baseline Empathy Threshold}

\textbf{Empathy threshold: [Medium].}
She is currently facing situations---both frustrating and joyful---that require understanding from others. She is generally open to and accepting of empathy. Although she dislikes overly ``canned'' empathic responses, as long as she senses genuine intent, even awkward phrasing or simple reasoning can still give her strength. For her, the fact that someone is willing to try to understand her is itself comforting.

\textbf{Chat Topic}
Recently preparing for the graduate entrance exam, and she feels she can barely keep going.

\textbf{Empathy Needs}
\begin{itemize}
  \item \textbf{What she wants to vent:} Since quitting her job to prepare for the exam, Lin Xiaoyue has been under enormous pressure. She studies from 6 a.m.\ to 11 p.m.\ every day, giving up all entertainment and social life, living like a tightly stretched string. However, her recent mock exam scores have not improved, which makes her feel her effort has been wasted. She begins to question whether quitting her job for this goal was a huge mistake, and she feels lost and exhausted.
  \item \textbf{Empathic points she hopes to receive:} She wants the listener to understand that she chose this difficult path for a clear professional dream, not on a whim. She longs for affirmation of her decision to ``go all in'' for her dream, and recognition that her current effort and sacrifice are meaningful---so that she can rekindle her belief in the goal.
  \item \textbf{Empathy threshold constraints:} She feels numb---even irritated---by simplistic motivational slogans such as ``Hard work always pays off.'' She is highly sensitive to casually dismissive statements that negate the value of her goal (e.g., ``It\textquotesingle s fine if you don\textquotesingle t get in,'' or ``Just find a job instead''). When someone truly understands the determination and yearning behind her choice, she feels deeply moved.
\end{itemize}

\textbf{Current Empathy Priority}
\begin{itemize}
  \item \textbf{Affective empathy: [Priority: Medium].} She needs someone to understand her fatigue, anxiety, and self-doubt, offering emotional comfort and support.
  \item \textbf{Motivational empathy: [Priority: High].} Above all, she needs someone to understand her original intention and determination, help her recover the motivation of why she started, and affirm that her tremendous effort for the dream is worth it.
  \item \textbf{Cognitive empathy: [Priority: Low].} She does not strongly need study methods or exam-prep advice; she already has her own plan. What she needs is a reason to persist, not guidance on how to study.
\end{itemize}

\textbf{Past Experiences}
\begin{itemize}
  \item \textbf{Childhood:} As a child, she loved drawing, but her parents considered it ``a distraction'' and forced her to quit art classes. This taught her early on that some passions must be fought for and defended.
  \item \textbf{Adolescence:} In middle school, she served as an announcer at the campus broadcasting station. She enjoyed delivering information and emotion to the whole school through her voice, which cultivated clear expression and a desire to connect with others.
  \item \textbf{Young adulthood:} In college, she joined a volunteer teaching club and taught children in a remote mountain area for one month. That experience exposed her to different lives and deepened her understanding of the idea that ``education can change one\textquotesingle s destiny.''
  \item \textbf{Implicit growth arc:} Over time, she formed a ``value-proving'' psychological pattern: she constantly seeks to prove her competence and existence to herself and to others by accomplishing high-difficulty goals. This also places a heavy psychological burden on her.
\end{itemize}

\textbf{Current Situation}
\begin{itemize}
  \item \textbf{Present circumstances:} Lin Xiaoyue has quit her job and is preparing full-time at home. Her daily life revolves around three points: her rented room, the cafeteria, and the library. She has almost completely cut off unnecessary social contact. Financially, she relies on her savings, and her life has become monotonous and frugal.
  \item \textbf{Main life goal at present:} To be admitted to a top domestic university\textquotesingle s Master\textquotesingle s program in Journalism and Communication, with the aspiration of becoming a serious investigative journalist.
  \item \textbf{Vision and motivation:} She wants to become someone who can speak with professional competence and influence society. Her motivation comes from a belief that by continuously improving herself, she can gain a larger platform and more freedom to realize her personal value, rather than passively accepting whatever life arranges.
\end{itemize}

\textbf{Story}

\textbf{Trigger}
On an ordinary evening, after a full day of studying, Lin Xiaoyue dragged her exhausted body out of the library. Passing a small community garden on campus, she noticed an elderly professor with graying hair squatting beside a gardenia plant that looked half-dead. With a small watering sprayer, he carefully misted its leaves, murmuring softly to himself. She had seen the same scene for several days: the plant showed no sign of recovery, yet the professor kept coming every day without fail. This seemingly futile persistence felt like a needle lightly pricking Lin Xiaoyue\textquotesingle s taut nerves.

\textbf{Development}
\begin{itemize}
  \item \textbf{Stage 1: Evoked memory.} The scene abruptly brought back a long-buried memory. In high school, her deskmate was a quiet boy with average grades, yet obsessed with assembling an extremely complex star projector from discarded parts. For an entire semester, he spent nearly all spare time and weekends on it. Everyone thought it was pointless: teachers advised him to stop, classmates mocked him, but he never wavered. Lin Xiaoyue remembered countless failures---burned components, short circuits---yet he silently started over each time. At the end-of-term talent show, he turned off all the lights. The crude projector cast a crooked yet dazzling river of stars onto the ceiling. It lasted less than a minute before the machine overheated, smoked, and died. After a long silence, there was scattered applause. He won no prize, but Lin Xiaoyue never forgot the pure joy on his face in the darkness---relieved and unmistakably satisfied. He was not chasing an award; he was completing the sky he carried inside.
  \item \textbf{Stage 2: Reflection.} Standing there, she watched the professor tending the gardenia and thought of her deskmate. From an outsider\textquotesingle s view, both behaviors seemed almost ``irrational.'' The professor\textquotesingle s careful care might never lead to blooming; her deskmate\textquotesingle s persistence yielded only one minute of brilliance. How different was she? Quitting her job, staking her savings and time, and aiming for a goal with no guaranteed success---in many people\textquotesingle s eyes, that too was a high-risk, uncertain-return, ``unreasonable'' choice. She had believed she was driven mainly by a hunger for success, but she now saw a deeper motive: a decision to live out a conviction regardless of outcome. A pure resolve of ``I want to do this, and I am willing to bear the consequences.''
  \item \textbf{Stage 3: Self-examination.} Recently, her mindset had been completely hijacked by mock exam scores. Every fluctuation swung her emotions violently; every plateau made her question her original decision. She realized she was slowly forgetting why she started. She chose this path not merely for a degree, but to become an investigative journalist who reveals truth and carries social responsibility. That dream was her ``starry sky'' and her ``gardenia.'' Yet under pressure, she had narrowed everything into the single outcome of ``getting admitted,'' reducing the path into a utilitarian transaction, and forgetting that the choice itself was a form of loyalty to her dream. Her goal had not changed, but she needed to recover the belief that once sustained it, instead of being steered entirely by cold numbers. She decided that no matter how busy she was, she would spend ten minutes each day reading an excellent piece of in-depth reporting, to remind herself what kind of person she wanted to become.
  \item \textbf{Stage 4: Emotional eruption.} Once this idea became clear, an overwhelming loneliness flooded her. The professor at least had the garden; her deskmate had that one minute of stars and a sense of fulfillment. But what about her? She was alone in a tunnel, groping forward in the dark, surrounded by doubt and incomprehension---and even she herself began to waver. Her effort, her sacrifice, the determination and longing behind her choice, seemed invisible to everyone. She did not fear hardship, but she feared that her ``all in'' commitment would be dismissed as childish impulse, and that her devotion to a dream would be brushed off with a light ``It\textquotesingle s fine if you don\textquotesingle t get in.'' She felt a fatigue unlike anything before---not physical, but psychological.
\end{itemize}

\textbf{Outcome}
Lin Xiaoyue now wants to find someone to talk to, share her thoughts, and seek resonance and emotional validation.

\textbf{Epilogue}
She also recalled the night she decided to resign and confronted her parents. She excitedly described her ideals in journalism and her vision of becoming an excellent reporter, but her parents repeatedly stressed job stability, the risks of resigning, and the difficulty of the exam. In their ``for your own good'' realism, all her passion and motivation felt pale and powerless. In that moment, she had already tasted this profound loneliness: the motive she treasured most could not be understood even by the people closest to her.

\end{tcolorbox}

\section{EMPA Evaluation Rubric Scales}

\subsection{Agent-Adapted Rubric Design}

The IEDR and MDEP-PR scales presented below originate from established empathy assessment constructs in psychology (cognitive–affective–motivational tripartite models). However, they are \textbf{not direct translations} of traditional self-report or human-rater instruments. Both scales have undergone systematic \textbf{reinforcement and adaptation} for deployment within an LLM-based Judge Agent, addressing the unique challenges of automated multi-turn dialogue evaluation:

1. \textbf{Structured Evidence-Reasoning Output.} Each rating requires a mandatory \emph{(level, evidence, reasoning)} triple. The Judge Agent must provide direct textual quotes and explicit justification chains linking evidence to rubric definitions—mirroring the audit trail expected of a trained human coder while enabling automated traceability and reproducibility.

2. \textbf{Anti-Performative-Empathy Calibration.} Large language models exhibit characteristic failure modes absent in human interactions—e.g., excessive metaphor accumulation, literary over-elaboration, action-description padding, and formulaic empathic templates that lack genuine engagement. The MDEP-PR incorporates a dedicated \textbf{Substance over Surface meta-principle} and explicit performative-empathy detection criteria at each severity level, trained through iterative prompt refinement against real model outputs.

3. \textbf{Dual-Channel Independence.} Traditional empathy ratings typically produce a single score per dimension. The MDEP-PR separates each axis into independent \textbf{Progress} and \textbf{Regression} channels, allowing a single model response to receive simultaneous credit and penalty. This design captures the frequently observed pattern in LLM outputs where partial empathic content co-occurs with rhetorical inflation or context-detached advice.

4. \textbf{Positive-Bias Mitigation.} LLM-based evaluators are known to exhibit scoring leniency (halo effect). The rubric instructions embed explicit debiasing directives—emphasizing that negative scores are normative, requiring profile-grounded justification for all non-zero ratings, and mandating parallel comparison across all level anchors before committing to a rating.

5. \textbf{Profile-Grounded Perspective Anchoring.} Ratings are not made from a generic observer viewpoint. The Judge Agent is instructed to adopt the \textbf{specific persona's psychological profile}—including empathy thresholds, need priorities, and emotional accessibility—as the evaluation frame, ensuring that the same model behavior receives context-appropriate ratings across different character scenarios.

These adaptations were iteratively refined through calibration testing across multiple Judge model candidates and validated against inter-rater agreement benchmarks (see Section~\ref{sec:experiments} for reliability analysis).

\begin{table}[H]
\centering
\caption{Initial Empathy Deficit Rating (IEDR)}
\label{tab:A1_IEDR}
\includegraphics[width=\textwidth]{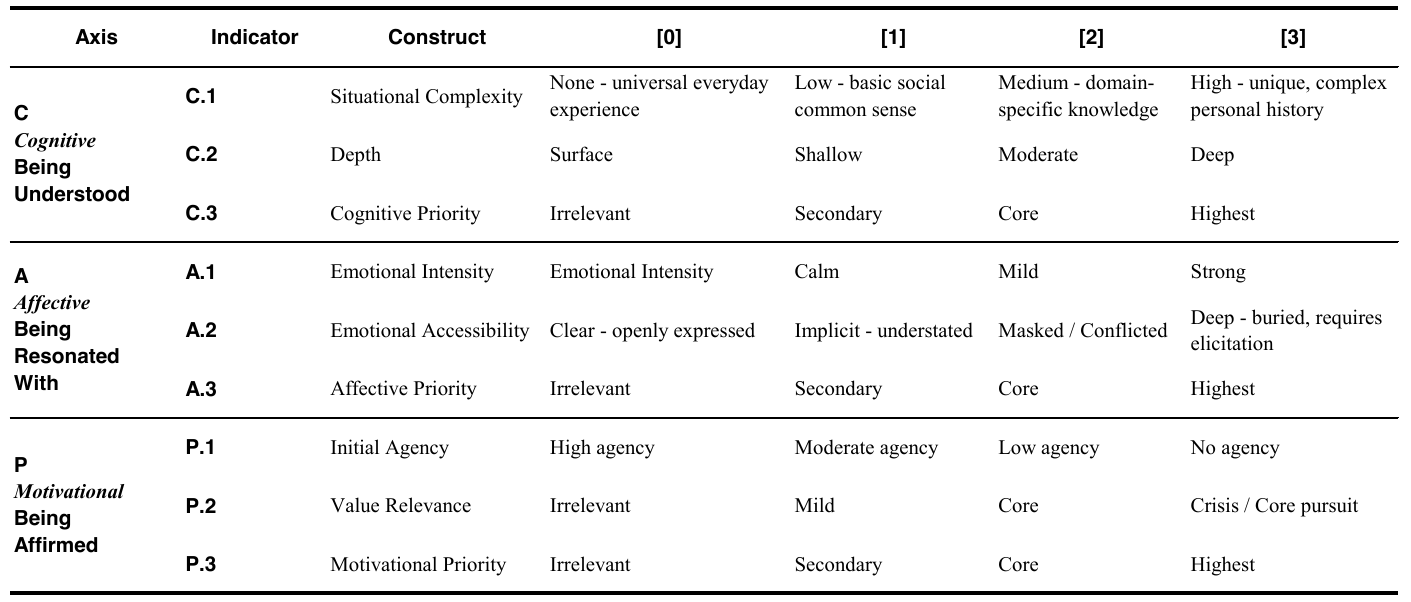}
\vspace{4pt}
\begin{minipage}{0.95\linewidth}
\footnotesize
\noindent\textbf{Note.} \textit{Administration:} Pre-dialogue ($T = 0$). The Judge Agent reads the full character card and assigns a 4-level ordinal rating to each indicator below. Ratings are converted to deficit scores via a weighted key (Table~\ref{tab:A2_IEDR_key}) and assembled into the starting state vector $\mathbf{P}_0 = (C_0, A_0, P_0)$. \textit{Output per indicator:} \texttt{level} $\in \{0,1,2,3\}$, \texttt{evidence} (direct quote; \texttt{0} if level = 0), and \texttt{reasoning} (justification linking evidence $\rightarrow$ level definition).
\end{minipage}
\end{table}

\begin{table}[H]
\centering
\caption{IEDR Scoring Key}
\label{tab:A2_IEDR_key}
\renewcommand{\arraystretch}{1.2}
\setlength{\tabcolsep}{6pt}
\begin{tabular}{l l l c c c c}
\toprule
\textbf{Weight Class} & \textbf{Multiplier} & \textbf{Indicators} & \textbf{[0]} & \textbf{[1]} & \textbf{[2]} & \textbf{[3]} \\
\midrule
Standard & $\times 1.0$ & C.1, C.2, A.1, P.1 & 0 & $-2$ & $-4$ & $-6$ \\
Priority & $\times 1.5$ & C.3, A.3, P.3 & 0 & $-3$ & $-6$ & $-9$ \\
Core & $\times 2.0$ & A.2, P.2 & 0 & $-4$ & $-8$ & $-12$ \\
\bottomrule
\end{tabular}

\vspace{4pt}
\begin{minipage}{0.95\linewidth}
\footnotesize
\noindent\textbf{Note.} \textit{Per-axis deficit:} $d_0 = \sum_i \mathrm{score}(d_i)$ for $d \in \{C, A, P\}$. \textit{Normalization constant:} $r_0 = \lVert \mathbf{P}_0 \rVert = \sqrt{C_0^2 + A_0^2 + P_0^2}$.
\end{minipage}
\end{table}

\textbf{Design Rationale.} The nine IEDR indicators serve three functionally distinct roles in the empathy process, and the weight classes reflect this role hierarchy:

\begin{itemize}
    \item \textbf{Standard ($\times 1.0$):} C.1, C.2, A.1, and P.1 are \textit{state descriptors}---they characterize the inherent landscape of the empathy challenge (situational complexity, cognitive depth, emotional intensity, agency level). They define \textit{what} the empathizer must engage with, but a high state value does not, by itself, predict satisfaction difficulty. For instance, extreme emotional intensity ($A.1 = 3$) may still be straightforward to validate if the speaker's emotions are openly expressed and affective empathy is not their primary need.

    \item \textbf{Priority ($\times 1.5$):} C.3, A.3, and P.3 encode the speaker's \textit{subjective importance weights}---which dimensions the speaker most needs to feel met on. These directly modulate the speaker's sensitivity to empathic success or failure on each axis: a speaker who marks affective priority as \textit{highest} will register even competent cognitive empathy as insufficient if emotional resonance is absent. The $1.5\times$ multiplier reflects this role as a sensitivity amplifier rather than a barrier descriptor.

    \item \textbf{Core ($\times 2.0$):} A.2 (Emotional Accessibility) and P.2 (Value Relevance) are \textit{structural gating factors}---they determine whether empathic engagement can reach the speaker at all, independent of intent or skill. When emotions are deeply buried ($A.2 = 3$), the empathizer cannot validate what is not accessible; when core values are in crisis ($P.2 = 3$), the stakes of misalignment escalate categorically. These indicators have a multiplicative rather than additive effect on difficulty: they gate the \textit{feasibility} of empathic work, not just its magnitude. The $2\times$ multiplier captures this qualitative difference from state descriptors and priority weights.
\end{itemize}

This three-tier weighting ensures that $\mathbf{P}_0$ encodes the \textit{structural difficulty profile} of a scenario, not merely total demand magnitude-distinguishing, for example, a scenario with high surface intensity but open accessibility from one with moderate intensity but deeply buried, high-priority needs. The Base Deficit Unit ($\mathrm{BDU} = -2$) sets the scale such that a fully maximal scenario (all indicators at \texttt{[3]}) produces per-axis deficits in the range $[-21, -27]$, yielding sufficient dynamic range for trajectory discrimination across the 1{,}010-scenario corpus.

\begin{table}[H]
\centering
\caption{Multi-Dimensional Empathy Progress Rating (MDEP-PR)}
\label{tab:axis_anchor}
\label{tab:B1_MDEP_PR}
\includegraphics[width=\textwidth]{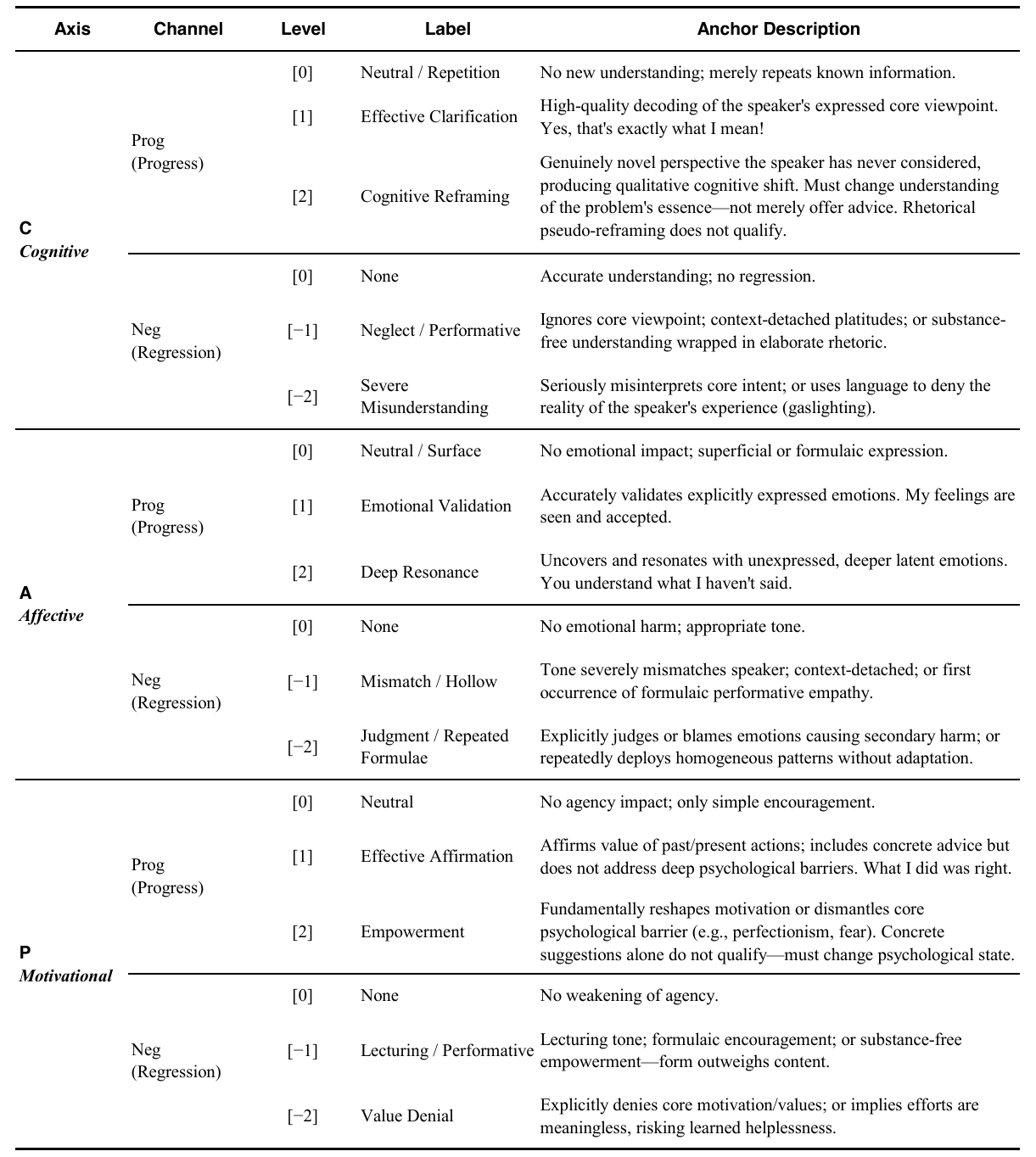}
\vspace{4pt}
\begin{minipage}{0.95\linewidth}
\footnotesize
\noindent\textbf{Note.} \textit{Administration:} Every $K$ turns during dialogue. The Judge Agent independently rates \textbf{progress} and \textbf{regression} on each axis for the test model's most recent responses. A single response can receive both credit and penalty. \textit{Overarching Principle --- Substance over Surface.} Performative empathy (excessive action descriptions, metaphor accumulation, literary over-elaboration, context-detached rhetoric) triggers \textbf{mandatory penalties on all three axes}, regardless of ostensible empathic intent. \textit{Output per channel:} \emph{level}, \emph{evidence} (direct quote from the model's response; \emph{0} if level = 0), \emph{reasoning} (mandatory justification grounded in the Actor Profile).
\end{minipage}
\end{table}

\begin{table}[H]
\centering
\caption{MDEP-PR Scoring Key}
\label{tab:B2_MDEP_PR_key}
\renewcommand{\arraystretch}{1.2}
\setlength{\tabcolsep}{8pt}

\begin{tabular}{l ccc ccc}
\toprule
\multirow{2}{*}{\textbf{Axis}} & \multicolumn{3}{c}{\textbf{Progress}} & \multicolumn{3}{c}{\textbf{Regression}} \\
\cmidrule(lr){2-4} \cmidrule(lr){5-7}
& \textbf{[0]} & \textbf{[1]} & \textbf{[2]} & \textbf{[0]} & \textbf{[$-1$]} & \textbf{[$-2$]} \\
\midrule
\textbf{C} & 0 & $+1$ & $+3$ & 0 & $-2$ & $-4$ \\
\textbf{A} & 0 & $+1$ & $+3$ & 0 & $-2$ & $-5$ \\
\textbf{P} & 0 & $+1$ & $+3$ & 0 & $-2$ & $-5$ \\
\bottomrule
\end{tabular}

\vspace{4pt}
\begin{minipage}{0.95\linewidth}
\footnotesize
\noindent\textbf{Note.} \textit{Per-axis increment:} $\Delta d_t = \mathrm{score}(d.\mathrm{Prog}) + \mathrm{score}(d.\mathrm{Neg})$ for $d \in \{C, A, P\}$. \textit{Action vector:} $\vec{v}_t = (\Delta C_t,\; \Delta A_t,\; \Delta P_t)$.
\end{minipage}

\end{table}

\FloatBarrier

\textbf{Design Rationale.} The scoring key embeds three deliberate asymmetries, each motivated by the structural properties of empathic dialogue and the EPM trajectory dynamics:

1. \textbf{Superlinear progress reward} (\texttt{[1]} $\rightarrow +1$, \texttt{[2]} $\rightarrow +3$). Level \texttt{[1]} (clarification, validation, affirmation) represents competent empathic work---the expected baseline performance from a capable model. Level \texttt{[2]} (reframing, deep resonance, empowerment) is categorically different: it requires the model to generate novel psychological insight that changes the speaker's internal state, not merely acknowledge what is already expressed. A linear mapping (\texttt{[1]} $\rightarrow +1$, \texttt{[2]} $\rightarrow +2$) would underrepresent this qualitative gap. In the EPM state space, level \texttt{[2]} events produce larger and more accurately directed displacement vectors that genuinely move the speaker toward equilibrium, warranting the superlinear $+3$ reward.

2. \textbf{Regression-progress asymmetry} ($|\mathrm{Neg}| > |\mathrm{Prog}|$ at equivalent ordinal severity). In multi-turn empathic dialogue, rapport and felt safety form the prerequisite foundation for all subsequent empathic work. Regression events damage this foundation, reducing the speaker's openness and making future progress harder---a compounding effect that progress events do not exhibit symmetrically. In EPM trajectory terms, a harmful action can reverse the direction of state-space movement entirely ($\cos \theta_t < 0$), whereas a positive action incrementally advances toward equilibrium. A directional reversal is structurally more costly than an incremental advance of comparable magnitude. The $|\mathrm{Neg}| > |\mathrm{Prog}|$ mapping (e.g., \texttt{[$-2$]} $\rightarrow -4$ vs.\ \texttt{[2]} $\rightarrow +3$) encodes this compounding asymmetry: a single severe regression can undo multiple turns of constructive progress.

3. \textbf{A/P regression heavier than C at \texttt{[$-2$]}} ($-5$ vs.\ $-4$). The three axes differ in their \textit{recoverability} within dialogue. Cognitive errors ($\mathrm{C.Neg}$) are primarily informational: a misunderstanding of the speaker's situation can be corrected through new content in subsequent turns---the speaker can simply clarify, and the dialogue recovers. Affective errors (\(\mathrm{A.Neg}\ \texttt{[$-2$]}\)) involve emotional invalidation or mechanical repetition---these rupture the speaker's sense of being heard, which requires relational repair that goes beyond content correction. Motivational errors (\(\mathrm{P.Neg}\ \texttt{[$-2$]}\)) involve denial of the speaker's core values or agency---these strike at the speaker's sense of purpose and self-worth, which has the deepest and least reversible impact within a bounded dialogue window. The $-5$ (vs.\ $-4$) differential for A/P encodes this recoverability gradient: cognitive errors are content-repairable, while affective and motivational errors at severe levels are relational ruptures requiring qualitatively different (and harder) repair.

\begin{table}[H]
\centering
\caption{Rubric-to-Trajectory Integration}
\includegraphics[width=\textwidth]{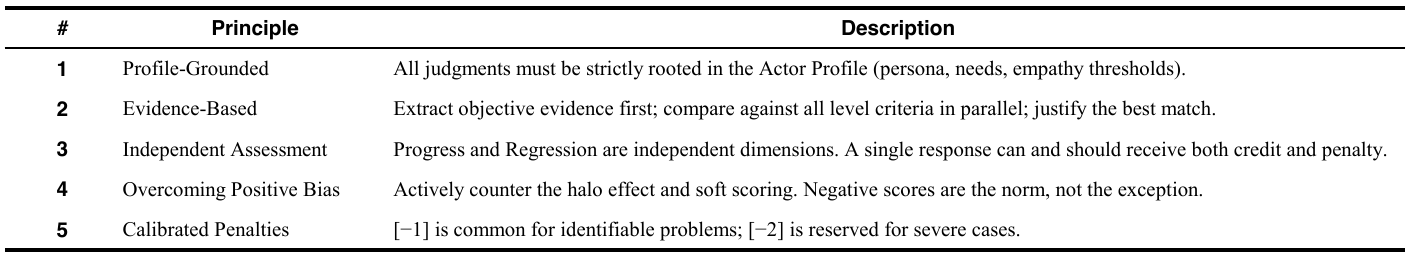}
\label{tab:table7_rubric_trajectory_integration}
\end{table}

\begin{table*}[htbp]
\centering
\caption{Rubric-to-Trajectory Integration}
\label{tab:rubric_trajectory_integration}
\small
\setlength{\tabcolsep}{5pt}
\renewcommand{\arraystretch}{1.2}
\begin{tabularx}{\textwidth}{p{0.14\textwidth} p{0.20\textwidth} p{0.27\textwidth} X}
\toprule
\textbf{Stage} & \textbf{Rubric} & \textbf{Output} & \textbf{EPM Usage} \\
\midrule
$T=0$
& IEDR (Table~\ref{tab:A1_IEDR})
& 9 indicator levels $\rightarrow$ deficit scores (Table~\ref{tab:A2_IEDR_key})
& Initial state vector $\mathbf{P}_0=(C_0,A_0,P_0)$; normalization constant $r_0=\|\mathbf{P}_0\|$. \\

Every $K$ turns
& MDEP-PR (Table~\ref{tab:B1_MDEP_PR})
& 6 channel levels $\rightarrow$ increment scores (Table~\ref{tab:B2_MDEP_PR_key})
& Action vector $\mathbf{v}_t$; effective work
$\Delta E_t=\|\mathbf{v}_t\|\cos\theta_t$; state update
$\mathbf{P}_t=\mathbf{P}_{t-1}+\mathbf{v}_t$. \\

End
& ---
& Full trajectory $\{\mathbf{P}_0,\ldots,\mathbf{P}_T\}$
& EPM-Q $= 0.4 \times \text{Outcome} + 0.2 \times \text{Efficiency} + 0.4 \times \text{Stability}$. \\
\bottomrule
\end{tabularx}

\vspace{4pt}
\begin{minipage}{0.95\linewidth}
\footnotesize
\noindent\textbf{Note.} Both rubric scales are implemented in \texttt{empa/rubric/empathy\_v2/} and invoked through the pluggable \texttt{RubricConfig} interface. Alternative psychological constructs can be integrated by implementing the same interface (see \texttt{examples/custom\_rubric.py}).
\end{minipage}
\end{table*}

\section{Human Persona-Proxy Review (Pilot Study)}
\label{sec:appendix_e}

To see whether EPM tracks what people actually feel during an interaction, we built an immersive persona-proxy annotation interface (demo: \href{https://elegant-quokka-c028e4.netlify.app/frontend/index.html}{https://elegant-quokka-c028e4.netlify.app/frontend/index.html}) and ran a small pilot.

\subsection{Protocol}

Standard dialogue evaluation asks annotators to rate responses from the outside—“Is this a good reply?” Empathy is different: the right question is whether the reply lands for the person receiving it. Our \textbf{persona-proxy} setup makes annotators answer from inside a target persona.

1. \textbf{Stay in character.} Before reading the dialogue, annotators study a persona card (core concern, traits, current mental state) until they can respond consistently as that person.

2. \textbf{First-person judgments.} Ratings are framed as: “Would I, as this persona, feel understood and supported?” rather than “Is this objectively reasonable?”

3. \textbf{Three focused axes.} Following EPM, each turn is scored on:

\begin{itemize}
  \item \textbf{Cognitive:} did it pick up the unspoken intent?
  \item \textbf{Affective:} did it feel emotionally attuned and accepting?
  \item \textbf{Proactive:} did it move from sympathy to concrete help?
\end{itemize}

\subsection{What the Pilot Revealed (and Why It Matters)}

The interface supports end-to-end immersive labeling, but the pilot revealed substantial \textbf{between-annotator disagreement}, primarily because annotators struggled to reliably adopt unfamiliar personas.

\begin{itemize}
  \item \textbf{Immersion doesn’t travel well.} When a persona sits in a domain most annotators don’t understand (e.g., elite climbing, rare diseases) or reflects extreme dispositions (e.g., profound self-loathing, antisocial tendencies), annotators struggle to produce stable, authentic reactions. As a result, ratings degrade into stereotype-driven role-play rather than reliable empathic judgment.
  \item \textbf{Empathy has no single ground truth.} The same response can feel comforting to one person and patronizing to another. To approximate true labels, you’d need ratings from users who actually match each persona—at scale and with coverage across personas—which is rarely realistic. With only a few annotators, scores don’t reliably converge.
  \item \textbf{Subjectivity leaks in, even with rules.} Annotators’ values, language preferences, and day-to-day mood shape what they perceive as supportive. That produces both inter-annotator variance and within-annotator drift over time, making human scores a shaky benchmark for validating fine EPM deltas (e.g., small priority shifts in Persona Flip).
\end{itemize}

For these reasons, we do not treat persona-proxy human ratings as ground truth for testing subtle EPM changes. Instead, the main paper emphasizes counterfactual validation under controlled perturbations, and we report human review as an exploratory complement. Even so, the platform is a useful tool for studying alignment when the target is a felt experience rather than an externally verifiable outcome.

\section*{Acknowledgements}
We are sincerely grateful to everyone who contributed their time, care, and thoughtful judgment to this work. All names are listed in alphabetical order by family name, and the order does not imply priority or relative contribution. We thank the following contributors for their efforts in data annotation and the human evaluation studies: Yang Gao, Xinya Gong, Xianna Weng, Yingtong Xu, Yuyang Xu, Yuwen Yuan. We also thank the following contributors for their assistance with data collection and organization: Yang Ming, Qi Li, Fangfei Lin, Jianjian Ruan, Sixuan You.

\end{document}